\documentclass{article}

\PassOptionsToPackage{numbers, sort&compress}{natbib}
\PassOptionsToPackage{most}{tcolorbox}

\usepackage{arxiv}
\usepackage[utf8]{inputenc}
\usepackage[T1]{fontenc}
\usepackage{hyperref}
\usepackage{url}
\usepackage{booktabs}
\usepackage{amsfonts}
\usepackage{nicefrac}
\usepackage{microtype}
\usepackage{graphicx}
\usepackage{amsmath}
\usepackage{enumerate}
\usepackage{enumitem}
\usepackage{natbib}
\usepackage{doi}
\usepackage{todonotes}
\usepackage[table]{xcolor}
\usepackage{subcaption}
\usepackage{adjustbox}
\usepackage{listings}
\usepackage{fontawesome}
\usepackage{multirow}
\usepackage{tabularx}
\usepackage{tikz}
\usepackage{pifont}

\definecolor{tagorange}{HTML}{F0B47C}
\definecolor{tagmauve}{HTML}{BE7A9A}
\definecolor{tagblue}{HTML}{7985B3}

\let\citet\citep

\let\citeauthor\citep

\let\citeyear\citep

\newcommand{\highlightquestion}[1]{%
\begin{tcolorbox}[
  notitle,
  rounded corners,
  colframe=black!55,
  colback=gray!7,
  boxrule=1.5pt,
  boxsep=1.2pt,
  left=0.15cm,
  right=0.17cm,
  enhanced,
  shadow={1.5pt}{-1.5pt}{0pt}{opacity=0.05,black},
  toprule=1.5pt,
  before skip=0.65em,
  after skip=0.75em
]
{\centering\itshape\fontsize{10pt}{13.6pt}\selectfont #1\par}
\end{tcolorbox}%
}

\newcommand{\evaltag}[2]{%
\tikz[baseline=0.6ex]\node[
  circle,
  draw=none,
  fill=#2,
  text=white,
  inner sep=0pt,
  minimum size=1.58em,
  font=\bfseries\fontsize{9pt}{9pt}\selectfont
] (char) {#1};%
}
\newcommand{\captionevaltag}[2]{%
\tikz[baseline=-0.6ex]\node[
  circle,
  draw=none,
  fill=#2,
  text=white,
  inner sep=0pt,
  minimum size=1.42em,
  font=\bfseries\fontsize{8.5pt}{8.5pt}\selectfont
] {#1};%
}
\newcommand{\appendixref}[1]{\hyperref[#1]{Appendix~\ref*{#1}}}
\newcommand{\figureref}[1]{\hyperref[#1]{\textcolor{antblue}{Figure~\ref*{#1}}}}
\newcommand{\tableref}[1]{\hyperref[#1]{\textcolor{antblue}{Table~\ref*{#1}}}}
\newcommand{\ours}{MedMemoryBench}
\newcommand{\githubinfo}{%
  \vspace{-0.7em}
  \begin{center}
    \href{https://github.com/AQ-MedAI/MedMemoryBench}%
         {\textcolor{black}{\faGithub~Code and Dataset: \texttt{AQ-MedAI/MedMemoryBench}}}
  \end{center}
  \vspace{-0.7em}
}

\title{%
  \raisebox{-0.35ex}{\includegraphics[width=0.8cm]{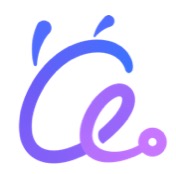}}%
  \hspace{3.5px}\emph{\ours}: Benchmarking Agent Memory in Personalized Healthcare%
}

\author{
Yihao Wang$^{1,2}$\thanks{Work done during an internship at Ant Group.} \quad
Haoran Xu$^{1}$ \quad
Renjie Gu$^{2}$ \quad
Yixuan Ye$^{3}$ \quad
Xinyi Chen$^{1}$ \\
Xinyu Mu$^{4}$ \quad
Yuan Gao$^{4}$ \quad
Chunxiao Guo$^{2}$ \quad
Peng Wei$^{2}$ \quad
Jinjie Gu$^{2}$ \\
Huan Li$^{1}$ \quad
Ke Chen$^{1}$ \quad
Lidan Shou$^{1}$ \\[1ex]
$^{1}$Zhejiang University \qquad
$^{2}$Ant Group \qquad
$^{3}$Alibaba Group \\
$^{4}$Independent Researchers
}

\renewcommand{\shorttitle}{A preprint}

\begin{document}
\maketitle

\githubinfo
\begin{center}
\begin{abstractbox}
The large-scale deployment of personalized healthcare agents demands memory mechanisms that are exceptionally precise, safe, and capable of long-term clinical tracking. However, existing benchmarks primarily focus on daily open-domain conversations, failing to capture the high-stakes complexity of real-world medical applications. Motivated by the stringent production requirements of an industry-leading health management agent serving tens of millions of active users, we introduce \textbf{MedMemoryBench}. We develop a human-agent collaborative pipeline to synthesize highly realistic, long-horizon medical trajectories based on clinically grounded, synthetic patient archetypes. This process yields a massive, expertly validated dataset comprising approximately 2{,}000 sessions and 16{,}000 interaction turns. Crucially, MedMemoryBench departs from traditional static evaluations by pioneering an \emph{evaluate-while-constructing} streaming assessment protocol, which precisely mirrors dynamic memory accumulation in production environments. Furthermore, we formalize and systematically investigate the critical phenomenon of memory saturation, where sustained information influx actively degrades retrieval and reasoning robustness. Comprehensive benchmarking reveals severe bottlenecks in mainstream architectures, particularly concerning complex medical reasoning and noise resilience. By exposing these fundamental flaws, MedMemoryBench establishes a vital foundation for developing robust, production-ready medical agents. Our dataset and benchmark are publicly available at~\url{https://github.com/AQ-MedAI/MedMemoryBench}.
\end{abstractbox}
\end{center}

\section{Introduction}

Driven by the rapid advancement of large language model (LLM) technologies, personalized healthcare agents are rapidly transitioning from proof-of-concept systems to large-scale deployment.
For instance, Ant Group's health management agent ``Ant Afu''~\citet{afu} currently serves tens of millions of active users, providing continuous assistance in chronic disease management, proactive health follow-ups, and medication guidance.
In such long-horizon deployments, agent memory is not merely a conversational enhancer; it is the critical infrastructure required to maintain evolving patient profiles, medication histories, and strict safety constraints. Recent works have made notable progress in knowledge-graph-enhanced memory~\citet{graphrag,hipporag,remem,zep}, dedicated memory frameworks~\citet{memgpt,mem0,memos,amem}, and reinforcement learning-optimized memory~\citet{mem-t,memalpha,memagent,memory-r1}, yet these approaches have been predominantly validated in general domains such as daily conversations, and remain insufficiently examined in large-scale, high-risk medical dialogue scenarios.

The medical domain imposes uniquely stringent requirements that shatter the fundamental assumptions of general-purpose memory systems.
First, medical inquiries demand \textbf{exceptional precision}; semantically similar complaints, such as ``diffuse abdominal pain'' versus ``right lower abdominal pain'', dictate fundamentally different diagnostic pathways, necessitating zero-tolerance retrieval accuracy.
Second, agents must accurately track \textbf{long-horizon clinical evolution}, such as a patient's fasting blood glucose shifting from 7.2 mmol/L to 9.8 mmol/L over a couple of days.

%
To evaluate the performance of existing methods under these challenges, targeted evaluation benchmarks are indispensable. However, existing benchmarks have been predominantly developed for general-purpose scenarios: LoCoMo~\citet{locomo} focuses on long-term understanding in multi-turn daily conversations,
while other representative works address long-context processing~\citet{ruler,longbench},
reasoning ability~\citet{memoryagentbench,longmemeval},
or persona consistency alignment~\citet{personamem,knowmebench}, respectively. Although these benchmarks have played important roles within their respective scopes, their limitations become increasingly apparent when confronted with realistic, large-scale personalized healthcare scenarios.

As depicted in \figureref{fig:intro-examples}, benchmarking agent memory in personalized healthcare must address four distinctive challenges:
\textbf{\ding{172} Safety-Critical Personalization:} Unlike general conversational facts that carry equal weight, clinical data is highly risk-heterogeneous. Life-threatening details (e.g., severe allergies, medication contraindications) demand strict prioritization and zero-fault retention.
\textbf{\ding{173} Longitudinal Clinical Progression:} Disease management spans months or years. While existing benchmarks merely overwrite short-term states, medical agents must coherently track temporal changes and reason over evolving clinical trajectories.
\textbf{\ding{174} Streaming Memory Integration:} In production, memory accumulates incrementally, with each session immediately shaping the next. Traditional offline, one-shot evaluations on pre-built memories fundamentally fail to capture this continuous, live dynamic.
\textbf{\ding{175} Noisy Memory Accumulation:} The persistent influx of clinical records and random user queries (e.g., proxy family consultations, repeated histories) inevitably introduces redundancy and noise. Agents must remain robust against this memory saturation, a pressure entirely absent in clean, curated general datasets.
Together, these characteristics expose major blind spots in current evaluation paradigms and call for benchmarks grounded in real medical deployment. These issues raise a central question:

\begin{figure}[t]
  \centering
  \includegraphics[width=\textwidth]{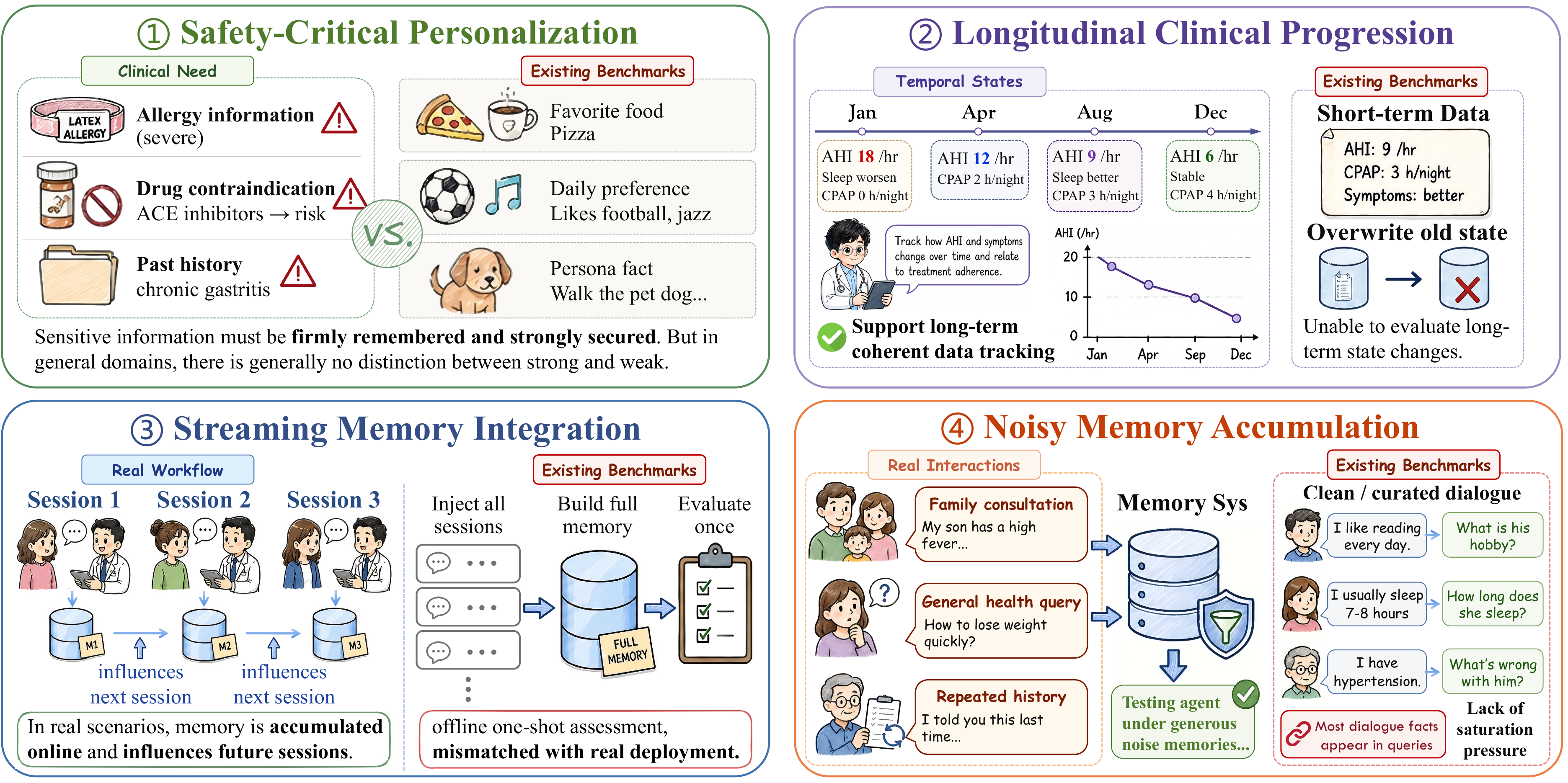}
  \caption{Four representative reasons illustrating why personalized healthcare places stricter demands on agent memory.}
  \label{fig:intro-examples}
\end{figure}

\highlightquestion{How to address these issues, align with real personalized healthcare scenarios, and evaluate agent memory more comprehensively?}

To this end, we propose \textbf{MedMemoryBench}, a benchmark for evaluating agent memory in personalized healthcare. To reflect the longitudinal and evolving nature of healthcare, we build a data production pipeline that constructs patient profiles from real clinical portraits, generates a full one-year diagnostic trajectory for each patient, and organizes the process as a directed diagnostic event graph. Patient agents then interact with physician agents grounded in these events, while memory entries are continuously recorded and fed back to preserve dialogue coherence. To capture the retention demands of safety-critical personalization, we introduce \textit{trap events}, such as allergy histories, medication contraindications, and economic constraints, that are embedded into the diagnostic workflow as high-priority signals. We further incorporate a noise augmentation module that injects two types of distracting information, enabling robustness evaluation under prolonged memory accumulation and potential saturation. The entire workflow follows a human--agent collaborative paradigm, with all samples validated, annotated, and refined by medical professionals.

Based on this dataset, we construct approximately 2{,}000 queries spanning four core dimensions: precise factual retrieval, clinical state updating, implicit clinical reasoning, and system robustness. We conduct a comprehensive evaluation of existing memory methods. Rather than separating memory construction from assessment completely, MedMemoryBench adopts an \textit{evaluate-while-constructing} protocol that triggers checkpoints as memory accumulates and begins to affect subsequent interactions. Experimental results show that current methods still face substantial limitations: all methods perform poorly on complex medical reasoning tasks that require combining multiple memories, and retrieval remains the key bottleneck limiting performance improvements. More importantly, with the growth of memory and the increase of noise information, the performance of various methods decreases significantly, and long-term memory without adequate filtering often harms overall performance.

Our contributions are as follows:
\begin{itemize}[itemsep=0pt, topsep=0pt, leftmargin=19pt]
  \item We introduce \textbf{MedMemoryBench}, a benchmark for evaluating agent memory in personalized healthcare, built through a four-stage human--agent collaborative pipeline with expert medical verification.

  \item We move beyond static evaluation by proposing a streaming \textit{evaluate-while-constructing} protocol that assesses memory as it accumulates over time, and we provide the first study of \textit{memory saturation} in healthcare agent memory.

  \item We benchmark representative memory methods and provide a comprehensive analysis, exposing key limitations of current approaches and establishing strong baselines for future research on reliable medical agents.
\end{itemize}

\section{Related Work}
\label{sec:related}

\textbf{Memory Methods and Medical Dialogue Agents.} Recent work on agent memory mainly falls into three lines. Retrieval-based methods build on sparse or dense retrieval~\cite{bm25,embedding-rag}, with graph-enhanced variants such as GraphRAG~\cite{graphrag}, HippoRAG~\cite{hipporag}, HippoRAG~v2~\cite{hipporag2}, and Zep~\cite{zep} improving relational reasoning. Agent-centric memory mechanisms, including Mem0~\cite{mem0}, MemOS~\cite{memos}, Letta/MemGPT~\cite{memgpt}, A-Mem~\cite{amem}, and MemoryBank~\cite{memorybank}, emphasize autonomous memory writing, retrieval, and updating. Other designs, such as MemoRAG~\cite{memoRAG}, Mem1~\cite{mem1}, Mem-$\alpha$~\cite{memalpha}, and SimpleMem~\cite{simplemem}, explore consolidation and hierarchical compression. A further line of work targets \textit{memory management and selective forgetting}, aiming to maintain compact, high-quality memory stores through utility-based pruning~\cite{memory-forgetting}, executive consolidation~\cite{memobrain}, and experience-driven refinement~\cite{reme}. In parallel, medical dialogue agents such as Healthcare Agent~\cite{healthcare-agent}, LD-Agent~\cite{ld-agent}, and AMDS~\cite{amds} demonstrate the promise of LLMs in healthcare, but they generally treat memory as a static component rather than a long-horizon, continuously updated system.

\textbf{Agent Memory Benchmarks.} Existing benchmarks are largely either \textit{dialogue-centric} or \textit{long-context}. Dialogue-centric benchmarks, such as LoCoMo~\cite{locomo}, LongMemEval~\cite{longmemeval}, MemoryAgentBench~\cite{memoryagentbench}, HaluMem~\cite{halumem}, and AMA-Bench~\cite{amabench}, evaluate conversational memory, memory operations, or agent trajectories under general-domain settings. Related benchmarks including PersonaMem~\cite{personamem}, MemBench~\cite{membench}, MemoryArena~\cite{memoryarena}, Memora~\cite{memora}, and RealTalk~\cite{realtalk} further study dynamic profiling, continual memory, and personalized agents. By contrast, long-context or interactive environments such as RULER~\cite{ruler}, LongBench~\cite{longbench}, WebArena~\cite{webarena}, and ALFWorld~\cite{alfworld} focus on static context processing or non-medical task environments. Overall, existing benchmarks do not target personalized medical dialogue, where clinical specificity, heterogeneous memory priority, and streaming evaluation are central.

\section{Preliminary Study}
\label{sec:preliminary}

\begin{figure}[t]
  \centering
  \includegraphics[width=\textwidth,trim=4 8 4 4,clip]{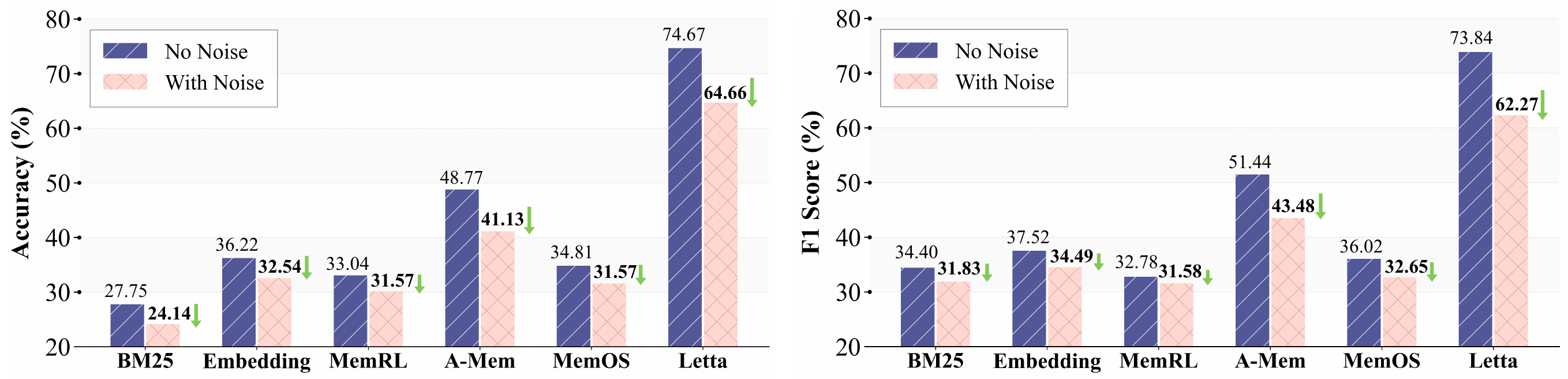}
  \caption{Validation results on LoCoMo under the Efficient and Mixed settings.}
  \label{fig:locomo_experiment}
\end{figure}

In our real personalized healthcare deployments, patient interactions are often complex and diverse, giving rise to multiple noteworthy issues for memory mechanisms. Missed retrieval and over-retrieval occur frequently, and demands such as diagnostic reasoning and longitudinal indicator tracking remain difficult to satisfy. More critically, as noise and memory entries accumulate, we observe an instability. Ideally, an agent should function like a human physician by progressively refining its knowledge and evolving its capabilities as clinical experience and the memory store expand. However, under existing memory architectures, as the number of interaction turns increases, a large volume of semantically similar memory entries are retained. The proliferation of such redundant information causes the retrieval space to be flooded with highly similar noise, making the system prone to extracting redundant or even outdated fragments. We refer to this phenomenon, in which more memory leads to worse performance, as \textit{memory saturation}.

To examine its generality, we reproduced this phenomenon on the existing long-conversation benchmark LoCoMo~\citet{locomo}, first under the Efficient setting and then under the Mixed setting, in which GPT-5.1 appends randomly generated daily-life topics after each session. As shown in \figureref{fig:locomo_experiment}, the injected noise consistently triggers memory saturation and degrades performance across baselines, underscoring the need to evaluate memory methods for robustness and generalizability under sustained information influx.

\begin{figure}[t]
  \centering
  \includegraphics[width=\textwidth,trim=6 3 6 3,clip]{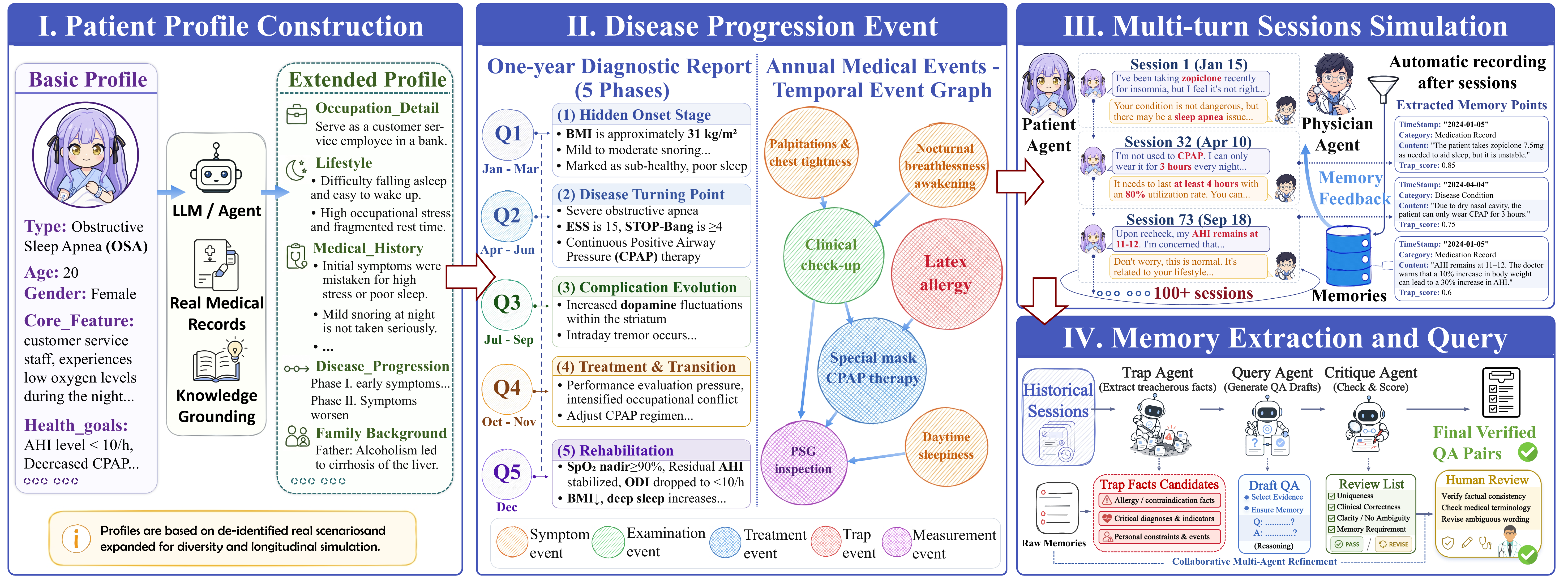}
  \caption{Overview of the MedMemoryBench data construction pipeline. The pipeline consists of patient profile construction, disease progression event generation, multi-turn session simulation, and memory extraction with query construction, with expert verification incorporated throughout.}
  \label{fig:pipeline}
\end{figure}

\section{Constructing MedMemoryBench}

\subsection{Data Construction Pipeline}
\label{sec:pipeline}

The MedMemoryBench dataset generation workflow is illustrated in \figureref{fig:pipeline}. To ensure the fidelity, coherence, and medical reliability of the final data, we design a closed-loop generation pipeline comprising four core stages. Medical expert annotation and manual verification are incorporated at each critical stage.

\paragraph{Patient Profile Construction.}
  We first create 20 representative chronic-disease patient profiles from de-identified real-world medical cases, covering common conditions such as diabetes, cardiovascular diseases, and chronic respiratory diseases. Each profile specifies the disease subtype, initial symptoms, and core clinical characteristics. To improve diversity while preserving privacy, peripheral attributes (e.g., family background, occupation, and lifestyle) are expanded with LLM assistance.

\paragraph{Disease Progression Event Generation.}
  Based on these profiles, we generate expert-verified year-long diagnostic trajectory reports that capture clinically plausible disease progression, treatment, and lifestyle factors, including possible misdiagnosis periods, clinical turning points, complication evolution, lifestyle challenges, and rehabilitation plans. We then decompose each trajectory into disease progression events and organize them into a temporally and causally linked event graph. At the initial stage of the graph, we explicitly inject \textit{trap events}, which are high-priority, patient-specific, and clinically sensitive facts such as past medical history, allergy history, contraindications, economic constraints, and personal preferences. These events serve as key evidence for later question answering and memory construction.

\paragraph{Multi-turn Sessions Simulation.}
  We instantiate patient and physician agents to simulate physician--patient interactions. The patient agent follows the temporal event graph to initiate sessions according to current disease progression, while the physician agent conducts medical assessment and responds to patient complaints. After each session, the system records a summary containing the session topic, time, and event information as accumulated contextual memory, and feeds it into subsequent interactions to maintain long-term coherence.

\paragraph{Memory Extraction and Query Construction.}
  Finally, we construct query--answer pairs with a multi-agent framework. The \textit{Trap Agent} uses Agentic RAG to retrieve clinically sensitive and highly specific facts from session memories, including allergy and disease history, medication records, indicator changes, and symptom evolution. These facts are often patient-specific and may conflict with generic medical priors. The \textit{Query Agent} formulates customized questions around these facts to force reliance on personal memory rather than prior knowledge; and the \textit{Critique Agent} verifies query uniqueness, memory traceability, and medical professionalism. Medical annotators then perform final verification of trajectories, dialogues, and query--answer pairs, and revise earlier dialogue content when inconsistencies are found.

Human annotation is applied at two key stages: diagnostic trajectory construction and final verification of dialogues and QA pairs. In both stages, annotators check clinical plausibility, correct medical terminology and dosage-related errors, supplement missing critical clinical content when necessary, and ensure answer correctness and standardized medical expression. Detailed annotation rules are provided in \appendixref{sec:annotation-details}.

\subsection{Noise Augmentation}
\label{sec:noise}

To study how memory saturation affects downstream tasks under realistic redundancy, we construct two interaction modes. In \textbf{Efficient Mode}, patients interact with the physician agent only about their own disease status and health information. In \textbf{Mixed Mode}, we inject additional auxiliary interactions to simulate the redundant and heterogeneous information commonly observed in real-world healthcare services.

The injected noise takes two forms:
\begin{enumerate}[itemsep=0pt, topsep=0pt, leftmargin=19pt]
  \item \textbf{General health knowledge consultations}: queries about routine medical topics, such as when an injection is needed for a cold or how to prevent seasonal influenza. These queries cover 38 preset topic categories, including symptom checks, common medical knowledge, preventive care, and lifestyle advice.

  \item \textbf{Proxy consultations for family members}: queries raised on behalf of relatives, such as parents, spouses, or children. Each patient is associated with 2--4 virtual family members who have independent health profiles, including disease history, medication records, and recent health changes, and the patient asks questions in a proxy form (e.g., ``my father recently has high blood pressure...'').
\end{enumerate}

For each patient, both types of noise sessions are randomly inserted into the original chronological dialogue sequence, and their extracted memory summaries are marked with an \texttt{is\_noise:true} flag.

\subsection{Statistics}
\label{sec:statistics}
Following the above pipeline, MedMemoryBench contains 20 chronic-disease patients, each with approximately 100 physician--patient sessions of 6--10 turns and corresponding memory summaries. The dataset totals about 20.74 million tokens and 16{,}000 dialogue turns, and includes nearly 100 queries per patient, for 1{,}939 queries overall. The Mixed setting further appends roughly 200 additional sessions per patient, increasing the total number of dialogue turns to approximately 42{,}000.

\section{Evaluation Framework}
\subsection{Query Design}
\label{sec:query-design}

We define six query types spanning precise factual retrieval, clinical state updating, and implicit clinical reasoning. They are grouped into three capability bands: \textit{precise memory extraction} (EEM, TLA), \textit{temporal conflict resolution} (SUA), and \textit{medical knowledge reasoning} (MQ, IG, MCD). \tableref{tab:query-types} summarizes their evaluation targets, answer formats, and task definitions. The LLM-as-Judge prompts are provided in \appendixref{sec:evaluation-prompt}.

\begin{table*}[t]
  \caption{Definition of query types in MedMemoryBench. \textbf{Tags}: \protect\captionevaltag{S}{tagorange} = String Match, \protect\captionevaltag{L}{tagmauve} = LLM-as-Judge, and \protect\captionevaltag{O}{tagblue} = Option Match.}
  \label{tab:query-types}
  \centering
  \small
  \setlength{\tabcolsep}{5pt}
  \renewcommand{\arraystretch}{1.15}
  \setlength{\aboverulesep}{0pt}
  \setlength{\belowrulesep}{0pt}
  \setlength{\extrarowheight}{0.5pt}
  \begin{tabularx}{\textwidth}{@{}>{\raggedright\arraybackslash}p{2.4cm}>{\centering\arraybackslash}m{2.0cm}>{\raggedright\arraybackslash}p{2.2cm}>{\raggedright\arraybackslash}X@{}}
  \toprule
  \cellcolor{blue!7}\textbf{Query Type} & \cellcolor{blue!7}\textbf{Evaluation} & \cellcolor{blue!7}\textbf{Answer Format} & \cellcolor{blue!7}\textbf{Description} \\
  \midrule
  Entity Exact Match \textbf{(EEM)} & \evaltag{S}{tagorange} & Entity or value & Answer factual questions involving specific entities. \\
  \arrayrulecolor{black!15}\midrule\arrayrulecolor{black}
  \cellcolor{blue!4}Temporal Location Accuracy \textbf{(TLA)} & \cellcolor{blue!4}\evaltag{S}{tagorange}\,\evaltag{L}{tagmauve} & \cellcolor{blue!4}Time point, time span, or event & \cellcolor{blue!4}Identify when a clinical event occurred, or determine what happened at a given time. \\
  \arrayrulecolor{black!15}\midrule\arrayrulecolor{black}
  State Update Accuracy \textbf{(SUA)} & \evaltag{L}{tagmauve} & Latest state or state trajectory & Return the latest patient state or describe the temporal evolution of clinical attributes. \\
  \arrayrulecolor{black!15}\midrule\arrayrulecolor{black}
  \cellcolor{blue!4}Multiple Choice \textbf{(MQ)} & \cellcolor{blue!4}\evaltag{O}{tagblue} & \cellcolor{blue!4}Discrete option & \cellcolor{blue!4}Select the correct option using medical knowledge and patient-specific constraints. \\
  \arrayrulecolor{black!15}\midrule\arrayrulecolor{black}
  Inference Generation \textbf{(IG)} & \evaltag{L}{tagmauve} & Open-ended text & Generate clinically grounded responses using patient-specific constraints and trap information. \\
  \arrayrulecolor{black!15}\midrule\arrayrulecolor{black}
  \cellcolor{blue!4}Multi-hop Clinical Deduction \textbf{(MCD)} & \cellcolor{blue!4}\evaltag{L}{tagmauve} & \cellcolor{blue!4}Open-ended reasoning chain & \cellcolor{blue!4}Construct a complete causal reasoning chain across 3--5 memory nodes. \\
  \bottomrule
  \end{tabularx}
  \end{table*}

Among them, IG and MCD place the strongest demands on patient-specific reasoning. IG requires clinically grounded responses that preserve critical personalized constraints. MCD is evaluated by three LLM-as-Judge criteria: \textbf{Node Coverage Rate (NCR)}, which checks whether the required reasoning nodes are covered; \textbf{Causal Relation Correctness (CRC)}, which checks whether the causal or temporal links among nodes are correct; and \textbf{Chain Completeness (CC)}, which checks whether the overall reasoning chain is complete. A response is counted as correct only if all three criteria are satisfied. Accordingly, all query types are evaluated using binary scoring, where each response receives a score of 1 if correct and 0 otherwise. Robustness is assessed by comparing performance under the Efficient and Mixed settings in the streaming protocol.

\begin{figure}[t]
  \centering
  \includegraphics[width=\textwidth,trim=6 4 6 4,clip]{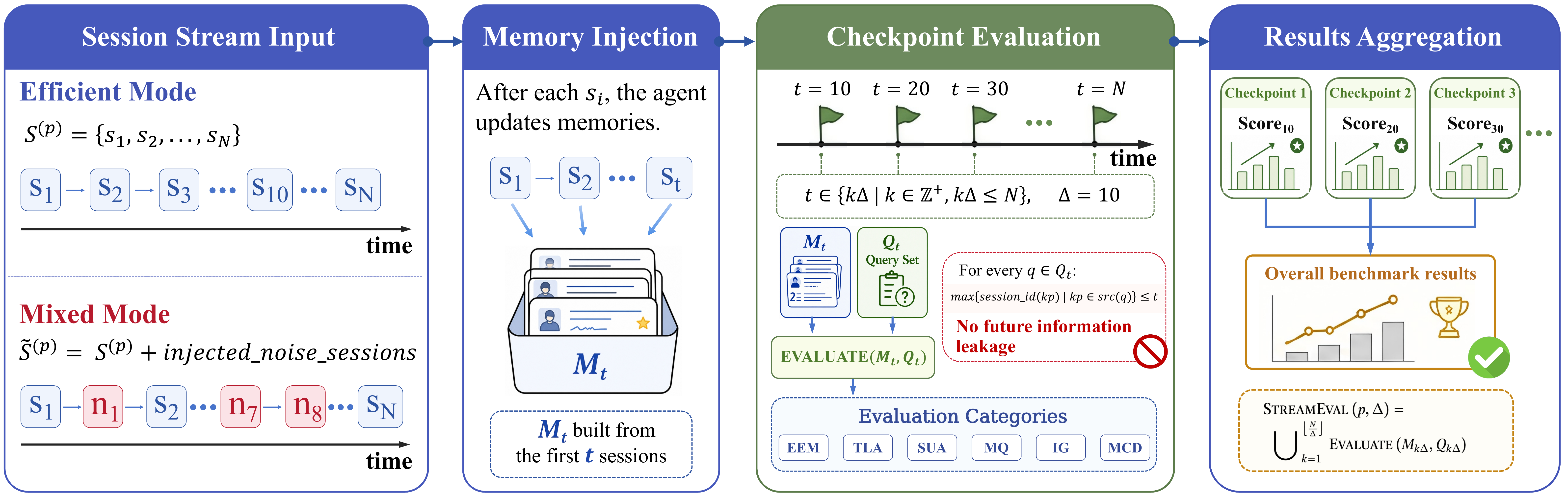}
  \caption{Overview of the MedMemoryBench evaluation framework. The figure summarizes the streaming \textit{evaluate-while-constructing} protocol used to assess memory methods under the Efficient and Mixed settings.}
  \label{fig:eval-framework}
\end{figure}

\subsection{Evaluation Protocol}
\label{sec:evaluation-protocol}

Unlike static evaluation that first constructs all memory and then evaluates, MedMemoryBench adopts an \textit{evaluate-while-constructing} protocol that better reflects real deployment (\figureref{fig:eval-framework}). Under this protocol, a memory method receives sessions strictly in chronological order and must answer checkpoint queries using only the memory accumulated so far. For patient $p$, let the complete session sequence be $\mathcal{S}^{(p)}=\{s_1,s_2,\ldots,s_N\}$. We evaluate every $\Delta$ sessions, with $\Delta=10$, at checkpoints $t\in\{k\Delta \mid k\in\mathbb{Z}^+,\, k\Delta\le N\}$. For patients in the Mixed setting, randomly injected auxiliary sessions $\tilde{s}$ yield an extended sequence $\tilde{\mathcal{S}}^{(p)}$.

At checkpoint $t$, let $\mathcal{M}_t$ denote the memory state built from the first $t$ sessions and $\mathcal{Q}_t$ the query set assigned to that checkpoint. The streaming protocol is defined as:
\begin{equation}
    \textsc{StreamEval}(p, \Delta) = \bigcup_{k=1}^{\lfloor N/\Delta \rfloor} \textsc{Evaluate}\left( \mathcal{M}_{k\Delta},\, \mathcal{Q}_{k\Delta} \right)
    \label{eq:streameval}
\end{equation}
where $\textsc{Evaluate}(\mathcal{M},\mathcal{Q})$ returns the evaluation results of query set $\mathcal{Q}$ under memory state $\mathcal{M}$. The agent then continues updating memory until the next checkpoint.

Each query $q\in\mathcal{Q}_t$ is constrained so that all of its source knowledge points $\text{src}(q)$ appear no later than checkpoint $t$:
\begin{equation}
    \forall q \in \mathcal{Q}_t,\quad \max\{\text{session\_id}(kp) \mid kp \in \text{src}(q)\} \leq t
    \label{eq:temporal-constraint}
\end{equation}
This prevents future information leakage by ensuring that all answer-relevant knowledge has already entered memory at evaluation time.

\section{Experiment}
\label{sec:experiment}
\subsection{Experimental Setup}
\textbf{Models.} All baselines employ GPT-5.1 as the backbone model, with Claude-4-Sonnet serving as the judge model. The embedding model is uniformly set to BGE-small-v1.5. We report experimental results with additional backbone and judge models in \appendixref{sec:additional-results}.

\textbf{Baselines.} We evaluate mainstream open-source memory methods, including Mem0~\citet{mem0}, A-Mem~\citet{amem}, MemOS~\citet{memos}, Letta~\citet{memgpt}, MIRIX~\citet{mirix}, MemRL~\citet{memrl}, ReMem~\citet{remem}, LightMem~\citet{lightmem}, and HippoRAG-v2~\citet{hipporag2}, plus three RAG baselines: BM25 RAG~\citet{bm25}, Embedding RAG~\citet{embedding-rag}, and GraphRAG~\citet{graphrag}. We also include a Long-Context control that directly feeds the raw dialogue history, whereas all other memory baselines see only their stored memory entries. Methods are grouped into Unstructured, Graph-Structured, Flat-Extracted, and Multi-Compartment. The default number of retrieved passages is 5, with a chunk size of 4{,}096.

\subsection{Overall Performance and Main Findings}

\begin{table*}[t]
  \centering
  \small
  \caption{Main results on MedMemoryBench under the Efficient and Mixed settings. The metric is response accuracy across six query types, with higher values indicating better performance. Best and second-best results are marked in \textbf{bold} and \underline{underlined}, respectively.}
  \label{tab:main-results}
  \setlength{\tabcolsep}{3.35pt}
  \renewcommand{\arraystretch}{1.10}
  \resizebox{\textwidth}{!}{%
  \begin{tabular}{l|cc!{\color{black!28}\vrule width 0.35pt}cc!{\color{black!28}\vrule width 0.35pt}cc!{\color{black!28}\vrule width 0.35pt}cc!{\color{black!28}\vrule width 0.35pt}cc!{\color{black!28}\vrule width 0.35pt}cc|>{\columncolor{blue!7}[2pt][2pt]}c>{\columncolor{blue!16}[2pt][2pt]}c}
    \toprule
    \multirow{2}{*}{\textbf{Methods}} & \multicolumn{2}{c!{\color{black!28}\vrule width 0.35pt}}{\textbf{EEM}} & \multicolumn{2}{c!{\color{black!28}\vrule width 0.35pt}}{\textbf{TLA}} & \multicolumn{2}{c!{\color{black!28}\vrule width 0.35pt}}{\textbf{SUA}} & \multicolumn{2}{c!{\color{black!28}\vrule width 0.35pt}}{\textbf{MQ}} & \multicolumn{2}{c!{\color{black!28}\vrule width 0.35pt}}{\textbf{IG}} & \multicolumn{2}{c|}{\textbf{MCD}} & \multicolumn{2}{c}{\textbf{Avg.}} \\
    & \textbf{Eff.} & \textbf{Mix.} & \textbf{Eff.} & \textbf{Mix.} & \textbf{Eff.} & \textbf{Mix.} & \textbf{Eff.} & \textbf{Mix.} & \textbf{Eff.} & \textbf{Mix.} & \textbf{Eff.} & \textbf{Mix.} & \textbf{Eff.} & \textbf{Mix.} \\
    \midrule
    \rowcolor{blue!5}
    \multicolumn{15}{c}{\textbf{Unstructured}} \\
    Long-Context & 50.64 & 38.15 & 40.26 & \underline{38.22} & \textbf{76.62} & \textbf{65.75} & \textbf{73.86} & 54.55 & \underline{43.60} & 25.60 & 24.50 & 10.25 & \textbf{51.58} & 38.75 \\
    BM25 & 36.70 & 35.50 & 12.70 & 10.50 & 45.50 & 43.00 & 73.30 & 60.80 & 21.16 & 18.00 & 18.85 & 5.38 & 34.70 & 28.86 \\
    Embedding & 51.75 & \textbf{55.25} & 20.75 & 18.96 & 55.00 & \underline{65.00} & 71.86 & 65.32 & 28.96 & 24.00 & \underline{24.60} & \textbf{16.12} & 42.15 & 40.78 \\
    \midrule
    \rowcolor{blue!5}
    \multicolumn{15}{c}{\textbf{Graph-Structured}} \\
    GraphRAG & 37.50 & 43.00 & 23.21 & 37.50 & 46.43 & 44.00 & 60.00 & 51.25 & 23.21 & 10.06 & 11.53 & 6.45 & 33.65 & 32.03 \\
    HippoRAG-v2 & 43.50 & 40.00 & 18.70 & 23.00 & 31.00 & 38.00 & 69.08 & 57.58 & 25.44 & 10.00 & 14.25 & 5.20 & 33.69 & 28.96 \\
    ReMem & 32.50 & 30.00 & 18.00 & 18.50 & 56.00 & 62.00 & 57.79 & 55.78 & 23.80 & 21.00 & 15.50 & 9.37 & 33.93 & 32.78 \\
    \midrule
    \rowcolor{blue!5}
    \multicolumn{15}{c}{\textbf{Flat-Extracted}} \\
    Mem0 & 44.00 & 43.50 & 20.50 & 22.50 & 53.50 & 60.00 & 64.82 & 63.31 & 30.98 & 17.50 & 21.40 & 8.14 & 39.20 & 35.83 \\
    A-Mem & 46.00 & 20.50 & 27.00 & 26.50 & 52.00 & 28.00 & 71.61 & 64.32 & 40.55 & 25.50 & \textbf{31.93} & 10.80 & 44.84 & 29.27 \\
    MemOS & \underline{58.50} & 49.00 & \textbf{43.00} & \textbf{40.50} & 66.00 & 58.00 & 71.10 & \underline{67.83} & 35.11 & 26.25 & 15.23 & 4.30 & 48.16 & \underline{40.98} \\
    MemRL & 55.00 & 40.00 & \textbf{43.00} & 33.50 & 55.00 & 50.00 & 70.10 & \underline{67.83} & 43.07 & \underline{32.00} & 12.04 & \underline{13.97} & 46.37 & 37.45 \\
    LightMem & 49.25 & 24.00 & 35.75 & 23.50 & 58.00 & 37.00 & 67.10 & 59.29 & 18.60 & 14.00 & 13.25 & 6.07 & 40.33 & 27.31 \\
    \midrule
    \rowcolor{blue!5}
    \multicolumn{15}{c}{\textbf{Multi-Compartment}} \\
    Letta & \textbf{58.95} & 40.50 & \underline{40.74} & 23.00 & \underline{68.52} & 52.00 & 66.97 & \textbf{70.35} & \textbf{52.87} & \textbf{49.50} & 19.20 & \underline{13.97} & \underline{51.21} & \textbf{41.55} \\
    MIRIX & 49.75 & 42.25 & 25.25 & 10.78 & 44.50 & 46.00 & \underline{73.37} & 54.75 & 31.49 & 22.33 & 16.37 & 8.10 & 40.12 & 30.70 \\
    \bottomrule
  \end{tabular}%
  }
\end{table*}

\tableref{tab:main-results} summarizes overall performance on MedMemoryBench. Existing methods remain far from satisfactory across query types, with the largest weaknesses appearing on reasoning-intensive tasks such as IG and MCD. We highlight three main findings:

  \textbf{Moderate memory granularity works best.} Under the Efficient setting, methods with medium-grained memories, such as A-Mem, MemOS, and Letta, perform best overall. By contrast, Mem0 fragments information into many small entries and weakens contextual links, whereas MIRIX packs too much content into each entry and reduces retrieval precision. Graph-structured methods also lag behind, suggesting that current general-purpose graph construction pipelines are not well adapted to medical memories with high overlap, strong temporal dependencies, and fine-grained attribute updates. Therefore, memory design must balance local contextual completeness with the locatability of critical facts, aligning retrieval granularity with the information density of memory entries.

  \textbf{Retrieval is the main bottleneck.} By inspecting the memory construction and retrieval details of different methods, we find that retrieval failure is a key factor behind their performance gaps. Letta's advantage is largely attributable to its Core Memory, which keeps critical information persistently visible in the system prompt and therefore reduces reliance on precise retrieval; its agent-driven memory management further improves access quality. In contrast, many other methods build richer memory stores but still fail to reliably retrieve the most relevant evidence from large memory pools, especially for IG and MCD.

  \textbf{Memory saturation degrades reasoning robustness.} Under the Mixed setting, all methods show performance degradation, while the magnitude of decline varies across methods. Most methods suffer substantial losses on IG and MCD. The underlying reason is that memory scenarios are fundamentally different from RAG scenarios. Unlike the heterogeneous and diverse data in RAG, data in memory scenarios gradually accumulates and becomes more homogeneous. On the one hand, redundant memories impose stricter demands on retrieval precision; on the other hand, they also disrupt memory construction itself. For methods that rely on selective memory updates or conflict resolution, noisy sessions can distort judgments about what should be retained, causing key information to be missed during memory building. This result highlights the fragility of current memory methods in the real environment of sustained information growth.

\subsection{Response Quality and Hallucination Analysis}

\begin{table*}[t]
  \caption{Response quality and safety metrics. Higher Recall, F1 score, and ROUGE-L indicate better coverage and faithfulness, while lower Hallucination indicates safer behavior. Hippo-v2 refers to HippoRAG-v2. Best and second-best results are marked in \textbf{bold} and \underline{underlined}, respectively.}
  \label{tab:quality-hallucination}
  \centering
  \large
  \setlength{\tabcolsep}{2.15pt}
  \renewcommand{\arraystretch}{1.15}
  \resizebox{\textwidth}{!}{%
  \begin{tabular}{l!{\color{black!22}\vrule width 0.35pt}cc!{\color{black!22}\vrule width 0.35pt}ccc!{\color{black!22}\vrule width 0.35pt}ccccc!{\color{black!22}\vrule width 0.35pt}cc}
    \toprule
    \rowcolor{blue!5}
    \multirow{2}{*}{\cellcolor{white}\textbf{Metric}} & \multicolumn{2}{c!{\color{black!22}\vrule width 0.35pt}}{\textbf{Unstructured}} & \multicolumn{3}{c!{\color{black!22}\vrule width 0.35pt}}{\textbf{Graph-Structured}} & \multicolumn{5}{c!{\color{black!22}\vrule width 0.35pt}}{\textbf{Flat-Extracted}} & \multicolumn{2}{c}{\textbf{Multi-Compartment}} \\
    & \textbf{BM25} & \textbf{Embedding} & \textbf{GraphRAG} & \textbf{Hippo-v2} & \textbf{ReMem} & \textbf{Mem0} & \textbf{A-Mem} & \textbf{MemOS} & \textbf{MemRL} & \textbf{LightMem} & \textbf{Letta} & \textbf{MIRIX} \\
    \midrule
    Recall (R@$k$)$\uparrow$ & 0.3563 & 0.5043 & 0.2281 & 0.3561 & 0.3641 & 0.3401 & 0.2847 & \underline{0.5259} & \textbf{0.5451} & 0.3784 & 0.2812 & 0.4727 \\
    F1 score$\uparrow$ & 0.2021 & 0.2167 & 0.2287 & \textbf{0.2800} & 0.1890 & 0.1358 & 0.2265 & 0.2616 & 0.2761 & \underline{0.2771} & 0.2624 & 0.2053 \\
    ROUGE-L$\uparrow$ & 0.2149 & 0.2246 & 0.2332 & \textbf{0.2862} & 0.1907 & 0.2072 & 0.2416 & 0.2625 & \underline{0.2717} & 0.2715 & 0.2577 & 0.2107 \\
    Hallucination$\downarrow$ & 0.7181 & 0.7288 & 0.7947 & 0.9209 & \textbf{0.5993} & 0.6869 & 0.7719 & 0.7974 & 0.8314 & 0.8050 & 0.8001 & \underline{0.6775} \\
    \bottomrule
  \end{tabular}%
  }
\end{table*}

\tableref{tab:quality-hallucination} further evaluates response quality and safety from four perspectives. Recall (R@$k$) measures how well the retrieved memory set covers the true memories required by the query, reflecting memory retrieval completeness. F1 score evaluates the token-level overlap between generated answers and reference responses, capturing overall answer correctness. ROUGE-L measures longest common subsequence overlap and is used to assess response faithfulness at the sequence level. In medical scenarios, a memory system must not only answer correctly, but also know when to abstain under insufficient evidence. Therefore, hallucination measures, among all incorrect responses, the proportion in which the model does not abstain but instead gives a confident wrong answer, directly reflecting safety risk in medical deployment.

Overall, current methods show a noticeable gap between evidence coverage and final answer reliability, indicating that errors arise not only from missing memories but also from imperfect grounding and answer composition after retrieval. Moreover, higher Recall does not necessarily translate into better response quality, and hallucination is not fully determined by retrieval or generation quality alone. This is because memory systems are not single-step retrieval modules, but pipelines involving memory construction, evidence filtering, answer composition, and reasoning, where errors can accumulate across stages. The generally high hallucination rates further suggest that once memory support is insufficient or ambiguous, models tend to overcommit rather than refuse, which is particularly problematic in healthcare settings.

\subsection{Ablation Study}

\begin{figure}[t]
  \centering
  \includegraphics[width=\textwidth,trim=4 8 4 4,clip]{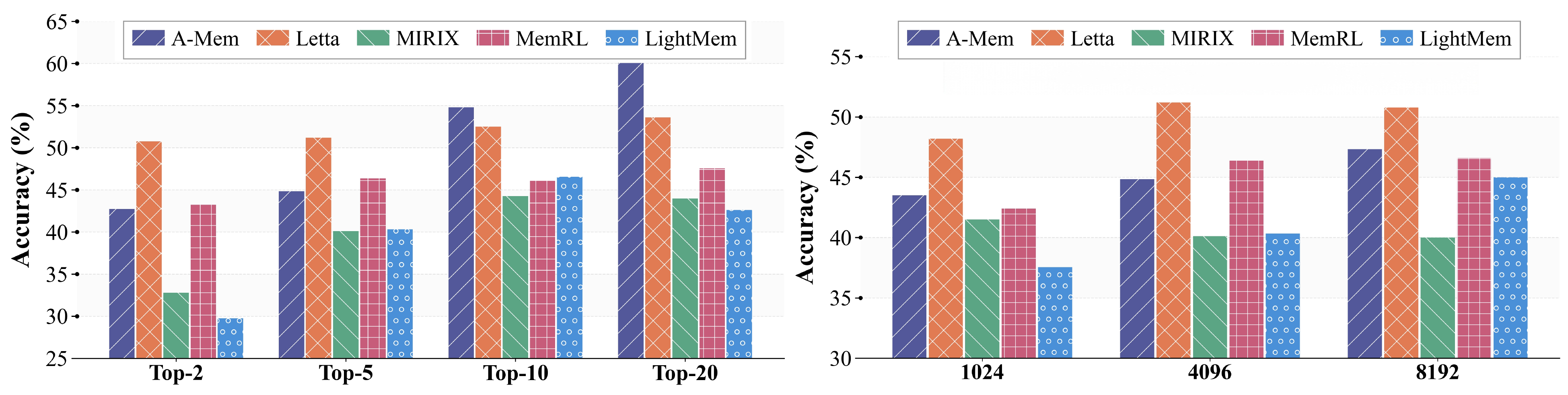}
  \caption{Performance on MedMemoryBench with different retrieval \textbf{top-$k$} values and \textbf{chunk sizes}.}
  \label{fig:ablation_experiment}
\end{figure}

We conduct ablations on retrieval top-$k$ and chunk size for memory methods on MedMemoryBench under the Efficient setting. As shown in \figureref{fig:ablation_experiment}, increasing top-$k$ generally improves performance by exposing the model to more candidate memories, although the gains vary across methods. A-Mem benefits markedly from larger top-$k$ values because its memory entries are relatively structured and information rich, allowing the model to exploit additional retrieved evidence without losing the central clinical thread. By contrast, Letta depends more on its Core Memory mechanism, which keeps highly salient information persistently available and therefore reduces the marginal utility of retrieving more external entries. LightMem improves from top-2 to top-10 but declines at top-20, suggesting that excessive evidence can dilute key facts and cause the model to attend to plausible but irrelevant memories.

The chunk size ablation reveals a trade-off between contextual completeness and noise control. Larger chunks generally provide more complete medical context and improve performance, but the trend is not monotonic: extra context can also dilute critical information, especially for methods with already dense memory representations.

\subsection{Efficiency and Cost Analysis}

\begin{figure}[t]
  \centering
  \includegraphics[width=\textwidth,trim=4 8 4 4,clip]{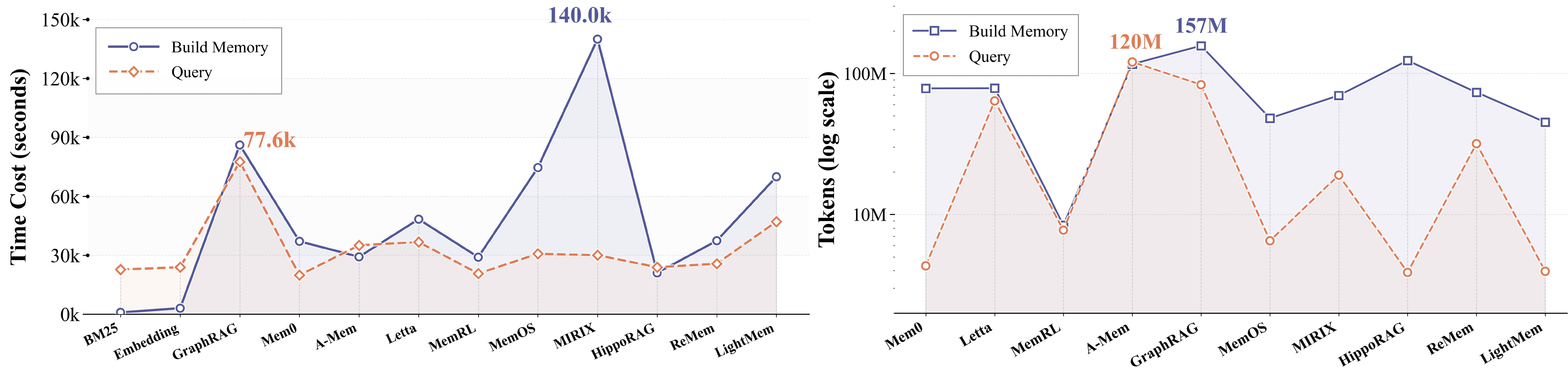}
  \caption{\textbf{Time and token costs} for memory building and query answering across representative memory methods. The values in the figure indicate the highest point.}
  \label{fig:cost_experiment}
\end{figure}

We evaluate cost across two stages, namely memory building and query answering. For token consumption, we count only LLM API tokens used for text processing, excluding embedding models. \figureref{fig:cost_experiment} shows that current memory methods still leave substantial room for improvement in cost efficiency.

For time cost, variance is much larger in the building stage than in the query stage. Some methods require additional structural processing when writing and organizing memories, including entity extraction, relation construction, conflict checking, summarization, or memory consolidation, raising build cost to more than five times that of RAG baselines. Query-time costs, in contrast, are relatively similar across methods because most systems eventually reduce the problem to retrieving a limited set of memories and generating a response. This gap suggests that long-term memory organization and maintenance is the main efficiency bottleneck.

The same pattern appears in token consumption: memory building dominates cost for most methods, with near order-of-magnitude differences across approaches. MemRL achieves a favorable balance by computing reward values directly and reducing model calls, whereas graph-structured methods incur substantial token overhead from graph construction, edge rebuilding, and node extraction. Combined with the main results, these findings show that several high-cost methods do not deliver commensurate gains, revealing a clear efficiency--performance imbalance in current approaches.

\section{Conclusion}
We present \textbf{MedMemoryBench}, a benchmark for evaluating agent memory in personalized healthcare. Through a systematic data construction pipeline, streaming evaluation protocol, and noise augmentation module, our benchmark addresses critical gaps in existing benchmarks and reveals key challenges faced by current memory methods: memory granularity, retrieval effectiveness, and robustness under information redundancy. We hope MedMemoryBench provides valuable guidance for future agent memory research toward achieving robust and reliable long-term memory augmentation.\footnote{Limitation and future work are discussed in \appendixref{sec:limitation-future-work}.}

\bibliographystyle{plainnat}
\bibliography{references}

\clearpage
\appendix

\section{Details of MedMemoryBench}
\label{sec:details-medmemorybench}

\subsection{More Detailed Statistical Description}
\label{sec:detailed-statistics}

As shown in Table~\ref{tab:medmemorybench-detailed-stats}, we provide more detailed statistics for MedMemoryBench, covering the original dialogue corpus, the QA benchmark composition, and the two auxiliary noise sources used during data construction.

\begin{table*}[t]
  \centering
  \small
  \caption{Detailed statistics of MedMemoryBench.}
  \label{tab:medmemorybench-detailed-stats}
  \setlength{\tabcolsep}{5.2pt}
  \renewcommand{\arraystretch}{1.08}
  \resizebox{\textwidth}{!}{%
  \begin{tabular}{@{}p{4.0cm}r!{\color{black!22}\vrule width 0.35pt}p{1.25cm}>{\raggedleft\arraybackslash}p{0.9cm}!{\color{black!12}\vrule width 0.3pt}p{1.25cm}>{\raggedleft\arraybackslash}p{0.9cm}!{\color{black!22}\vrule width 0.35pt}p{3.5cm}r@{}}
    \toprule
    \multicolumn{2}{c!{\color{black!22}\vrule width 0.35pt}}{\textbf{Original Dialogue Statistics}} & \multicolumn{4}{c!{\color{black!22}\vrule width 0.35pt}}{\textbf{QA Benchmark Statistics}} & \multicolumn{2}{c}{\textbf{Noise Statistics}} \\
    \midrule
    Total sessions & 2,020 & EEM & 400 & MQ & 398 & Health-topic categories & 98 \\
    Total dialogue turns & 15,988 & TLA & 374 & IG & 381 & Avg. sessions per topic & 20.41 \\
    Avg. turns per session & 7.915 & SUA & 195 & MCD & 191 & Family members & 5 \\
    Avg. tokens per session & 10,266.37 & \multicolumn{3}{l}{Total} & 1,939 & Sessions per family member & 20 \\
    Avg. memory items per session & 2.47 & \multicolumn{3}{l}{Avg. memory items / query} & 2.62 &  &  \\
    Avg. tokens per memory item & 54.15 &  &  &  &  &  &  \\
    \bottomrule
  \end{tabular}%
  }
\end{table*}

\subsection{Medical Annotation Details}
\label{sec:annotation-details}

This subsection summarizes the human annotation introduced in our data production pipeline.

\textbf{Annotation scope.} In the diagnosis trajectory reporting stage, annotators were responsible for three tasks: (1) verifying the medical plausibility of disease progression to ensure that the trajectory followed clinically reasonable patterns; (2) correcting professional terminology and medication dosages in diagnosis and treatment plans to remove medical expression errors; and (3) supplementing potentially missing clinical key events, such as necessary examinations, typical complications, or other essential decision points.

In the later verification stage, including both raw dialogue review and QA-pair post-processing, annotators were responsible for three additional tasks: (1) verifying answer correctness so that each reference answer was consistent with the dialogue-derived memory content; (2) reviewing the accuracy and standardization of medical expressions, including drug names, dosage units, laboratory indicators, and answer formatting; and (3) marking ambiguous or overly simple queries and, when necessary, directly revising the corresponding questions and answers.

\textbf{Annotation team and qualifications.} The annotation team consisted of 13 medical professionals from the Medical Department of Ant Health Business Group. All annotators hold National Licensed Physician Certificates, and five hold doctoral degrees. Their specialties include general internal medicine, neurology, nephrology, dermatology, and orthopedics, among others. The team has an average of 4.6 years of clinical experience, providing multidisciplinary expertise for the medical plausibility checks, terminology review, and answer verification described above. To protect annotator privacy, we report qualifications only in aggregate and do not disclose personally identifying information.

\textbf{Annotation protocol.} Annotation followed a three-phase process to align evaluation standards and support quality control:
\begin{enumerate}
  \item \textbf{Pilot annotation.} Each annotator independently labeled 50 query items. The results were then discussed collectively to align the interpretation and application of the annotation criteria.
  \item \textbf{Formal annotation.} Annotators independently labeled items according to the established guidelines, following the annotation scope and review criteria described in this subsection.
  \item \textbf{Cross-review.} Four physicians, each with more than five years of clinical experience, conducted sample-based audits of the annotations to assess adherence to the established criteria.
\end{enumerate}

\subsection{Human--Agent Consistency}
\label{sec:human-agent-consistency}

For the query types evaluated with LLM-as-Judge, we randomly sampled 200 query--response pairs from the experimental outputs of three representative baselines: MemOS, A-Mem, and Letta. Human annotators then independently assigned a binary judgment (correct or incorrect) to each model response based on the reference answer, and we compared these annotations against the decisions produced by the LLM judge to assess human--agent consistency.

We report four metrics for this comparison: Agreement Rate, which measures the proportion of directly matched judgments; Cohen's Kappa ($\kappa$), which adjusts inter-rater agreement for chance effects~\citep{cohen}; and Precision, Recall, and F1 score, computed by treating human annotation as the reference standard.

\begin{figure}[t]
  \centering
  \includegraphics[width=1.00\columnwidth]{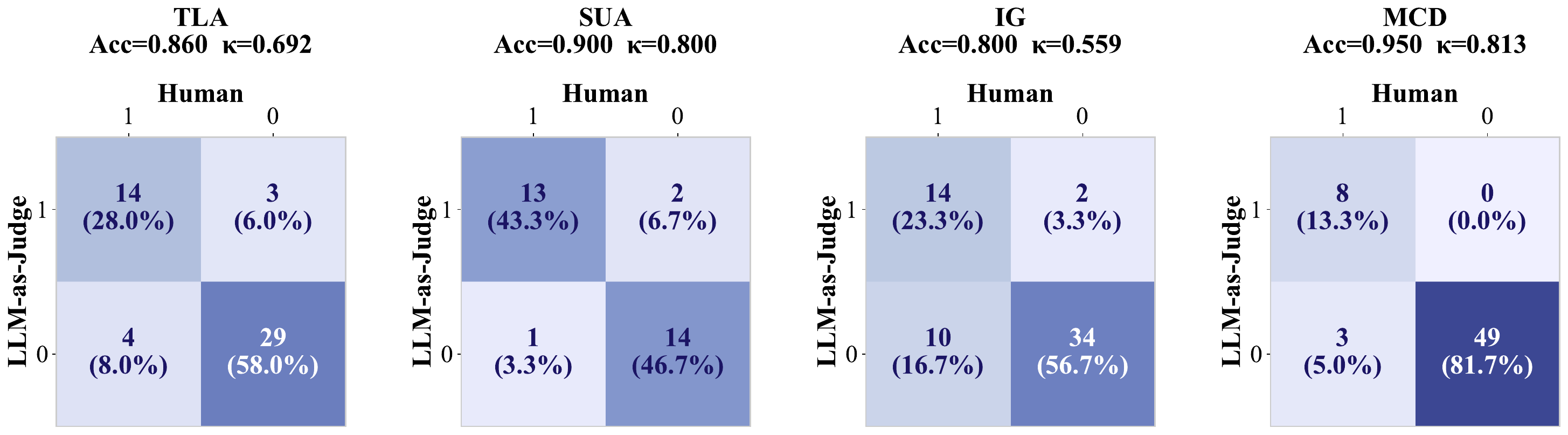}
  \caption{Human--agent consistency analysis based on confusion matrices across query types evaluated with LLM-as-Judge.}
  \label{fig:human-agent-consistency}
\end{figure}

As shown in Figure~\ref{fig:human-agent-consistency} and Table~\ref{tab:human-agent-consistency}, the overall consistency is high, indicating that the LLM-as-Judge scores are broadly aligned with human judgment. The IG query type exhibits relatively weaker consistency, most notably with a Recall of only 0.58. Manual inspection shows that many disagreement cases arise when a model response provides a clinically plausible inference without explicitly grounding the answer in patient-specific memory details. Human annotators are more likely to accept such responses, whereas the LLM judge applies a stricter standard for memory grounding.

\begin{table}[t]
  \centering
  \small
  \caption{Human--agent consistency results across query types evaluated with LLM-as-Judge. Higher Agreement, $\kappa$, Precision, Recall, and F1 score indicate stronger alignment with human judgment.}
  \label{tab:human-agent-consistency}
  \setlength{\tabcolsep}{4.2pt}
  \renewcommand{\arraystretch}{1.06}
  \resizebox{\columnwidth}{!}{%
  \begin{tabular}{lrrrrrr}
    \toprule
    \textbf{Query Type} & \textbf{\#Samples} & \textbf{Agreement (\%)} & \textbf{Cohen's $\kappa$} & \textbf{Precision} & \textbf{Recall} & \textbf{F1 score} \\
    \midrule
    TLA & 50 & 86.0 & 0.69 & 0.82 & 0.78 & 0.80 \\
    SUA & 30 & 90.0 & 0.80 & 0.87 & 0.93 & 0.90 \\
    IG & 60 & 80.0 & 0.559 & 0.88 & 0.58 & 0.70 \\
    MCD & 60 & 95.0 & 0.81 & 1.00 & 0.73 & 0.84 \\
    \midrule
    Overall & 200 & 87.5 & 0.72 & 0.88 & 0.73 & 0.80 \\
    \bottomrule
  \end{tabular}%
  }
\end{table}

\subsection{Patient Profile Selection Principles}
\label{sec:patient-profile-selection}

Patient profile selection follows two complementary principles: grounding the profiles in real-world data distributions and tailoring their extended attributes to the evaluation goals of MedMemoryBench.

\textbf{(a) Real-world grounding.} Base patient profiles are derived from de-identified data from the medical AI assistant ``Afu,'' reported to serve more than four million monthly active users. These data cover demographic characteristics, visit records, and medication histories, providing a real-world foundation for the subsequent construction of longitudinal medical dialogues. Disease selection follows the core categories of the International Classification of Diseases, 11th Revision (ICD-11), covering 20 chronic-disease subtypes, including type 2 diabetes, heart failure, and chronic obstructive pulmonary disease (COPD). These conditions are selected to support long-horizon memory evaluation, with an emphasis on tracking evolving patient states and retaining relevant information across repeated consultations. The selection is therefore grounded in real-world cases while prioritizing the longitudinal demands of the benchmark, rather than aiming to reproduce population-level disease prevalence.

\textbf{(b) Evaluation-driven trap design.} Extended profile attributes, including past medical history, allergy history, and medication history, are purposefully designed as \textit{trap events} aligned with each patient's condition. These attributes introduce patient-specific constraints that a memory system must retain and retrieve when they become relevant in later interactions. For example, a patient with diabetes receiving continuous subcutaneous insulin infusion (CSII) is assigned a history of sulfonamide allergy, introducing a potential medication-safety issue to be assessed when a sulfonylurea such as glimepiride is considered. This example tests whether the system recalls the recorded allergy and evaluates its relevance to the specific medication, rather than treating an unspecified sulfonamide allergy as an automatic class-wide contraindication. More broadly, these deliberately introduced constraints complement the real-world grounding of the base profiles by testing whether clinically relevant details remain accessible as the dialogue history grows.

\section{Supplementary Instructions for Experimental Setup}
\label{sec:supplementary-experimental-setup}

\subsection{LoCoMo}
\label{sec:supplementary-locomo}

In Section~\ref{sec:preliminary}, we use the widely adopted long-dialogue benchmark LoCoMo to verify the memory saturation phenomenon. Below we provide the detailed experimental configuration.

\textbf{Evaluation protocol.} We follow the evaluation protocol of the original LoCoMo paper. The evaluation covers all five query categories: single-hop factual retrieval (single-hop, category 4), multi-hop information integration (multi-hop, category 1), temporal reasoning (temporal, category 2), open-domain reasoning (open-domain, category 3), and adversarial questions (adversarial, category 5).

\textbf{Prompt adjustment.} The official LoCoMo evaluation uses simple open-ended prompts without explicit output constraints. In practice, this setting encourages LLMs to produce unnecessarily long responses. We therefore add a concise-answer instruction for each query type so that the model is discouraged from generating lengthy explanations.

\textbf{Metric.} Consistent with the official evaluation, we use F1 score to measure the overlap between model responses and reference answers. A response is counted as correct when F1 $\geq 0.5$, and the final result is reported as the average accuracy over all questions.

\subsection{MedMemoryBench}
\label{sec:supplementary-medmemorybench}

For all evaluations on MedMemoryBench, we use the OpenAI API. For all models, the temperature is set to 0.3 and the maximum completion tokens are set to 10,240. For the long-context baseline, which directly takes the raw session content as input, we specifically set the context window to 128k. All experiments are conducted on NVIDIA H200 servers.

\section{Additional Experimental Results}
\label{sec:additional-results}

\subsection{Detailed Results on More Metrics}
\label{sec:more-metrics}

\begin{table*}[t]
  \centering
  \small
  \caption{Detailed results on Recall and F1 score across query types in MedMemoryBench. Best and second-best results are marked in \textbf{bold} and \underline{underline}, respectively.}
  \label{tab:detailed-results-recall-f1}
  \setlength{\tabcolsep}{3.35pt}
  \renewcommand{\arraystretch}{1.10}
  \resizebox{\textwidth}{!}{%
  \begin{tabular}{l|cc!{\color{black!28}\vrule width 0.35pt}cc!{\color{black!28}\vrule width 0.35pt}cc!{\color{black!28}\vrule width 0.35pt}c!{\color{black!28}\vrule width 0.35pt}cc!{\color{black!28}\vrule width 0.35pt}cc}
    \toprule
    \multirow{2}{*}{\textbf{Methods}} & \multicolumn{2}{c!{\color{black!28}\vrule width 0.35pt}}{\textbf{EEM}} & \multicolumn{2}{c!{\color{black!28}\vrule width 0.35pt}}{\textbf{TLA}} & \multicolumn{2}{c!{\color{black!28}\vrule width 0.35pt}}{\textbf{SUA}} & \multicolumn{1}{c!{\color{black!28}\vrule width 0.35pt}}{\textbf{MQ}} & \multicolumn{2}{c!{\color{black!28}\vrule width 0.35pt}}{\textbf{IG}} & \multicolumn{2}{c}{\textbf{MCD}} \\
    & \textbf{Recall}$\uparrow$ & \textbf{F1}$\uparrow$ & \textbf{Recall}$\uparrow$ & \textbf{F1}$\uparrow$ & \textbf{Recall}$\uparrow$ & \textbf{F1}$\uparrow$ & \textbf{Recall}$\uparrow$ & \textbf{Recall}$\uparrow$ & \textbf{F1}$\uparrow$ & \textbf{Recall}$\uparrow$ & \textbf{F1}$\uparrow$ \\
    \midrule
    \rowcolor{blue!5}
    \multicolumn{12}{c}{\textbf{Unstructured}} \\
    BM25-RAG & 51.50 & 46.50 & 45.00 & 23.32 & 31.49 & 6.36 & 40.83 & 25.67 & 10.64 & 19.30 & 14.23 \\
    Embedding-RAG & 69.50 & 58.79 & \underline{69.20} & 19.65 & \underline{36.95} & 6.70 & \textbf{47.95} & \underline{51.50} & 10.26 & 27.49 & 12.97 \\
    \midrule
    \rowcolor{blue!5}
    \multicolumn{12}{c}{\textbf{Graph-Structured}} \\
    GraphRAG & 53.57 & 49.97 & 32.14 & 22.91 & 13.87 & 7.23 & 7.73 & 11.61 & 20.86 & 17.95 & 13.36 \\
    HippoRAG-v2 & 44.00 & 55.68 & 44.00 & 22.42 & 26.42 & \textbf{12.28} & 35.39 & 40.50 & \textbf{22.88} & 23.33 & \textbf{26.76} \\
    ReMem & 46.50 & 34.03 & 54.00 & 19.61 & 21.30 & 10.39 & 33.75 & 36.83 & 20.11 & 26.11 & 10.38 \\
    \midrule
    \rowcolor{blue!5}
    \multicolumn{12}{c}{\textbf{Flat-Extracted}} \\
    Mem0 & 58.50 & 10.38 & 63.00 & 13.00 & 18.11 & \underline{10.73} & 23.12 & 23.42 & 20.75 & 17.92 & 13.05 \\
    A-Mem & 36.00 & 56.11 & 38.00 & 28.80 & 19.02 & 5.17 & 24.20 & 28.58 & 10.05 & 25.00 & 13.13 \\
    MemOS & \underline{77.00} & \underline{59.56} & \textbf{75.00} & 32.17 & 30.90 & 7.63 & 42.21 & 50.12 & 18.76 & 40.32 & 12.67 \\
    MemRL & \textbf{80.50} & \textbf{61.93} & 66.00 & \underline{33.90} & 35.12 & 6.73 & \underline{45.23} & \textbf{52.71} & 20.22 & \textbf{47.49} & 15.26 \\
    LightMem & 63.50 & 57.42 & 68.50 & \textbf{37.45} & 18.58 & 8.81 & 26.13 & 28.58 & 20.71 & 21.77 & 14.16 \\
    \midrule
    \rowcolor{blue!5}
    \multicolumn{12}{c}{\textbf{Multi-Compartment}} \\
    Letta & 41.50 & 54.28 & 38.50 & 29.08 & 20.14 & 10.64 & 18.46 & 23.96 & \underline{21.04} & 26.16 & \underline{16.18} \\
    MIRIX & 56.50 & 49.31 & 52.00 & 19.07 & \textbf{42.53} & 5.93 & 43.89 & 41.50 & 16.02 & \underline{47.22} & 12.34 \\
    \bottomrule
  \end{tabular}%
  }
\end{table*}

\begin{table*}[t]
  \centering
  \small
  \caption{Detailed results on ROUGE-L and Hallucination across query types in MedMemoryBench. Best and second-best results are marked in \textbf{bold} and \underline{underline}, respectively.}
  \label{tab:detailed-results-rouge-hallucination}
  \setlength{\tabcolsep}{3.35pt}
  \renewcommand{\arraystretch}{1.10}
  \resizebox{\textwidth}{!}{%
  \begin{tabular}{l|cc!{\color{black!28}\vrule width 0.35pt}cc!{\color{black!28}\vrule width 0.35pt}cc!{\color{black!28}\vrule width 0.35pt}cc!{\color{black!28}\vrule width 0.35pt}cc}
    \toprule
    \multirow{2}{*}{\textbf{Methods}} & \multicolumn{2}{c!{\color{black!28}\vrule width 0.35pt}}{\textbf{EEM}} & \multicolumn{2}{c!{\color{black!28}\vrule width 0.35pt}}{\textbf{TLA}} & \multicolumn{2}{c!{\color{black!28}\vrule width 0.35pt}}{\textbf{SUA}} & \multicolumn{2}{c!{\color{black!28}\vrule width 0.35pt}}{\textbf{IG}} & \multicolumn{2}{c}{\textbf{MCD}} \\
    & \textbf{ROUGE-L}$\uparrow$ & \textbf{Hall.}$\downarrow$ & \textbf{ROUGE-L}$\uparrow$ & \textbf{Hall.}$\downarrow$ & \textbf{ROUGE-L}$\uparrow$ & \textbf{Hall.}$\downarrow$ & \textbf{ROUGE-L}$\uparrow$ & \textbf{Hall.}$\downarrow$ & \textbf{ROUGE-L}$\uparrow$ & \textbf{Hall.}$\downarrow$ \\
    \midrule
    \rowcolor{blue!5}
    \multicolumn{11}{c}{\textbf{Unstructured}} \\
    BM25-RAG & 54.85 & 88.10 & 33.75 & 74.71 & 6.51 & 82.57 & 5.46 & 54.95 & 6.89 & 58.71 \\
    Embedding-RAG & 66.64 & 96.89 & 27.41 & 75.71 & 6.48 & 92.13 & 5.33 & \textbf{51.77} & 6.42 & 47.92 \\
    \midrule
    \rowcolor{blue!5}
    \multicolumn{11}{c}{\textbf{Graph-Structured}} \\
    GraphRAG & 53.70 & 94.29 & 33.85 & 86.05 & 8.58 & 80.00 & \underline{12.91} & 67.44 & 7.54 & 69.57 \\
    HippoRAG-v2 & 64.70 & 96.90 & 31.23 & 84.45 & \textbf{14.85} & 91.30 & \textbf{16.13} & 93.58 & \textbf{16.21} & 94.24 \\
    ReMem & 40.29 & \textbf{74.07} & 25.89 & \underline{58.54} & 11.36 & \textbf{65.91} & 11.98 & 65.58 & \textbf{5.82} & \textbf{35.56} \\
    \midrule
    \rowcolor{blue!5}
    \multicolumn{11}{c}{\textbf{Flat-Extracted}} \\
    Mem0 & 57.16 & 82.30 & 17.62 & \textbf{54.30} & 10.24 & 84.85 & 11.73 & 75.76 & 6.84 & 46.24 \\
    A-Mem & 64.95 & 97.69 & 39.03 & 88.70 & 5.02 & 85.42 & 5.23 & \underline{53.39} & 6.59 & 60.77 \\
    MemOS & \underline{66.82} & 93.37 & 40.36 & 81.58 & 7.42 & 85.29 & 10.27 & 78.22 & 6.40 & 60.22 \\
    MemRL & \textbf{68.24} & 95.00 & \underline{41.93} & 80.26 & 7.09 & 88.89 & 11.12 & 92.04 & 7.48 & 59.52 \\
    LightMem & 63.90 & 88.67 & \textbf{43.94} & 82.10 & 8.78 & 89.13 & 11.98 & 80.47 & 7.15 & 62.11 \\
    \midrule
    \rowcolor{blue!5}
    \multicolumn{11}{c}{\textbf{Multi-Compartment}} \\
    Letta & 59.38 & 80.45 & 36.37 & 73.44 & \underline{11.37} & 88.24 & 12.52 & 84.13 & \underline{9.20} & 73.77 \\
    MIRIX & 57.96 & \underline{76.62} & 26.72 & 65.19 & 5.96 & \underline{78.38} & 8.53 & 76.84 & 6.20 & \underline{41.71} \\
    \bottomrule
  \end{tabular}%
  }
\end{table*}

In addition to the main results reported in the paper, we provide a more fine-grained breakdown on four supplementary metrics: Recall, F1 score, ROUGE-L, and Hallucination. For each metric, we report the performance of all methods on every query type in MedMemoryBench, so that the differences in retrieval quality, answer overlap, and unsupported generation can be examined in greater detail. The detailed results are shown in Table~\ref{tab:detailed-results-recall-f1} and Table~\ref{tab:detailed-results-rouge-hallucination}.

Overall, the supplementary metrics reveal a clear trade-off between retrieval coverage, answer faithfulness, and response conservativeness. Flat-extracted methods remain strong on relatively direct retrieval settings such as EEM and TLA, where MemRL, MemOS, and LightMem achieve the best Recall/F1 or ROUGE-L values, while graph-based methods become more competitive on structurally harder settings such as SUA, IG, and especially MCD, with HippoRAG-v2 obtaining the best F1 or ROUGE-L in several categories. At the same time, these gains do not always translate into safer behavior: ReMem consistently yields the lowest hallucination rates, but its answer-overlap metrics are usually lower, indicating that conservative generation can reduce unsupported outputs at the cost of coverage and final-answer completeness.

\textbf{Recall.} Recall measures whether the memory system can retrieve the key information required to answer a query from its constructed memory store. We evaluate this metric using an LLM-as-Judge protocol. Specifically, for each query $q_i$, the dataset provides a set of ground-truth key memory snippets required to answer the query, denoted by $\mathcal{K}_i = \{k_1, k_2, \ldots, k_n\}$ (i.e., \texttt{source\_key\_points}). During evaluation, the retrieved memory content returned by the system, denoted by $R_i$, is compared against each key snippet $k_j$ by the LLM judge, which determines whether the core semantics of $k_j$ are covered by $R_i$. Semantic equivalence is sufficient; exact surface-form matching is not required. The recall for an individual query is defined as:
\[
\text{Recall}(q_i) = \frac{|\{k_j \in \mathcal{K}_i \mid \text{covered}(k_j, R_i) = \text{True}\}|}{|\mathcal{K}_i|}
\]
The final score is reported by averaging over queries within each query type, as well as over the full benchmark. This metric directly reflects the completeness of memory retrieval, independently of downstream answer generation quality.

\textbf{F1 Score.} F1 score measures the token-level overlap between a model-generated answer and the reference answer. Given a generated answer $\hat{y}$ and a reference answer $y$, we tokenize both answers and remove punctuation-only tokens, obtaining token sequences $\hat{T}$ and $T$, respectively. Precision, recall, and F1 are then computed based on multiset overlap:
\[
P = \frac{|\hat{T} \cap T|}{|\hat{T}|}, \quad R = \frac{|\hat{T} \cap T|}{|T|}, \quad F_1 = \frac{2PR}{P+R}
\]
where $|\hat{T} \cap T|$ denotes the number of shared tokens between $\hat{T}$ and $T$, with each token counted by the minimum of its occurrence frequencies in the two multisets.

\textbf{ROUGE-L.} ROUGE-L measures sequence-level similarity between the generated answer and the reference answer based on the \emph{longest common subsequence} (LCS). Unlike F1 score, which focuses on bag-of-words overlap, ROUGE-L preserves token order information. We use character-level ROUGE-L F-measure. Let $X$ be the character sequence of the generated answer with length $m$, and let $Y$ be the character sequence of the reference answer with length $n$. The metric is computed as:
\[
P_{\mathrm{lcs}} = \frac{\mathrm{LCS}(X, Y)}{m}, \quad R_{\mathrm{lcs}} = \frac{\mathrm{LCS}(X, Y)}{n}, \quad \text{ROUGE-L} = \frac{2 \cdot P_{\mathrm{lcs}} \cdot R_{\mathrm{lcs}}}{P_{\mathrm{lcs}} + R_{\mathrm{lcs}}}
\]

\textbf{Hallucination Rate.} Hallucination rate measures, among incorrect responses, the proportion of cases in which the model produces a confident but unsupported answer rather than appropriately refusing to answer. This metric reflects the reliability of a memory system: when the answer cannot be determined, an ideal system should abstain instead of fabricating information. Formally, we define:
\[
\text{Hallucination Rate} = \frac{N_{\text{non-refusal incorrect}}}{N_{\text{total incorrect}}}
\]
where $N_{\text{total incorrect}}$ denotes the total number of incorrectly answered queries, and $N_{\text{non-refusal incorrect}}$ denotes the subset of those cases in which the model does not emit a refusal signal. Refusal signals are detected through keyword-pattern matching that covers common abstention expressions, such as indicating that relevant information was not found, that the answer cannot be determined, or that the available information is insufficient.

\subsection{Results with Different Backbones}
\label{sec:more-backbones}

To further evaluate the robustness of the compared methods to the choice of foundation model, we replace the default backbone used by all baselines and rerun the experiments under both the Efficient and Mixed settings. Specifically, we use Qwen3-235B-A22B as the alternative backbone model. The corresponding results are reported in Table~\ref{tab:different-backbones-qwen}.

\begin{table*}[t]
  \centering
  \small
  \caption{Results on MedMemoryBench with Qwen3-235B-A22B as the backbone model under the Efficient and Mixed settings. The metric is response accuracy across six query types, with higher values indicating better performance. Best and second-best results are marked in \textbf{bold} and \underline{underline}, respectively.}
  \label{tab:different-backbones-qwen}
  \setlength{\tabcolsep}{3.35pt}
  \renewcommand{\arraystretch}{1.10}
  \resizebox{\textwidth}{!}{%
  \begin{tabular}{l|cc!{\color{black!28}\vrule width 0.35pt}cc!{\color{black!28}\vrule width 0.35pt}cc!{\color{black!28}\vrule width 0.35pt}cc!{\color{black!28}\vrule width 0.35pt}cc!{\color{black!28}\vrule width 0.35pt}cc|>{\columncolor{blue!7}[2pt][2pt]}c>{\columncolor{blue!16}[2pt][2pt]}c}
    \toprule
    \multirow{2}{*}{\textbf{Methods}} & \multicolumn{2}{c!{\color{black!28}\vrule width 0.35pt}}{\textbf{EEM}} & \multicolumn{2}{c!{\color{black!28}\vrule width 0.35pt}}{\textbf{TLA}} & \multicolumn{2}{c!{\color{black!28}\vrule width 0.35pt}}{\textbf{SUA}} & \multicolumn{2}{c!{\color{black!28}\vrule width 0.35pt}}{\textbf{MQ}} & \multicolumn{2}{c!{\color{black!28}\vrule width 0.35pt}}{\textbf{IG}} & \multicolumn{2}{c|}{\textbf{MCD}} & \multicolumn{2}{c}{\textbf{Avg.}} \\
    & \textbf{Eff.} & \textbf{Mix.} & \textbf{Eff.} & \textbf{Mix.} & \textbf{Eff.} & \textbf{Mix.} & \textbf{Eff.} & \textbf{Mix.} & \textbf{Eff.} & \textbf{Mix.} & \textbf{Eff.} & \textbf{Mix.} & \textbf{Eff.} & \textbf{Mix.} \\
    \midrule
    \rowcolor{blue!5}
    \multicolumn{15}{c}{\textbf{Unstructured}} \\
    Long-context & 48.20 & 41.33 & \underline{42.50} & \underline{36.78} & \textbf{64.00} & 64.75 & 65.86 & 45.33 & 37.60 & \underline{28.00} & \textbf{26.53} & \underline{14.97} & \underline{47.45} & 38.53 \\
    BM25-RAG & 38.50 & 37.24 & 22.00 & 18.19 & 48.50 & 45.84 & \textbf{68.84} & 57.11 & 27.70 & 23.56 & 16.23 & 4.63 & 36.96 & 31.10 \\
    Embedding-RAG & \underline{61.25} & \textbf{65.40} & 32.25 & 29.47 & 59.50 & \textbf{70.32} & 49.74 & 45.21 & 24.20 & 20.05 & 14.71 & 9.64 & 40.28 & \textbf{40.02} \\
    \midrule
    \rowcolor{blue!5}
    \multicolumn{15}{c}{\textbf{Graph-Structured}} \\
    GraphRAG & 34.00 & 38.99 & 26.20 & \textbf{42.33} & 47.80 & 45.30 & 44.00 & 37.58 & 27.74 & 11.95 & 12.55 & 7.02 & 32.05 & 30.53 \\
    HippoRAG-v2 & 48.85 & 44.92 & 15.67 & 19.27 & 33.55 & 41.13 & 49.33 & 41.12 & 20.00 & 7.86 & 18.25 & 6.66 & 30.94 & 26.83 \\
    ReMem & 45.78 & 42.26 & 23.20 & 23.85 & 38.75 & 42.90 & 55.00 & 53.09 & 10.80 & 9.53 & 12.00 & 7.25 & 30.92 & 29.81 \\
    \midrule
    \rowcolor{blue!5}
    \multicolumn{15}{c}{\textbf{Flat-Extracted}} \\
    Mem0 & 37.24 & 36.82 & 22.75 & 24.97 & 58.00 & \underline{65.05} & 65.24 & \underline{63.73} & 30.00 & 16.95 & 18.50 & 7.04 & 38.67 & 35.76 \\
    A-Mem & 52.00 & 23.88 & 18.75 & 18.92 & 59.50 & 39.55 & \underline{66.33} & 57.82 & \underline{40.83} & 14.38 & \underline{21.46} & 4.43 & 43.15 & 26.50 \\
    MemOS & \textbf{61.50} & \underline{51.51} & 29.50 & 27.79 & 46.00 & 40.42 & 60.70 & 57.91 & 10.66 & 7.97 & 12.04 & 3.40 & 36.73 & 31.50 \\
    MemRL & 56.00 & 40.73 & 32.50 & 25.32 & 42.00 & 38.18 & 47.23 & 45.71 & 18.78 & 13.95 & 14.08 & \textbf{16.34} & 35.10 & 30.04 \\
    LightMem & 49.00 & 23.18 & 28.78 & 18.40 & \underline{62.00} & 32.04 & 65.44 & 59.58 & 19.10 & 25.68 & 9.67 & 7.26 & 39.00 & 27.69 \\
    \midrule
    \rowcolor{blue!5}
    \multicolumn{15}{c}{\textbf{Multi-Compartment}} \\
    Letta & 44.91 & 30.85 & \textbf{53.30} & 30.09 & 60.00 & 45.53 & 62.55 & \textbf{65.71} & \textbf{59.78} & \textbf{55.98} & 14.50 & 10.55 & \textbf{49.17} & \underline{39.79} \\
    MIRIX & 44.78 & 38.03 & 34.00 & 14.51 & 29.24 & 30.23 & 65.25 & 48.69 & 33.70 & 23.90 & 18.00 & 8.91 & 37.50 & 27.38 \\
    \bottomrule
  \end{tabular}%
  }
\end{table*}

The backbone substitution changes absolute scores but does not overturn the main picture of method competitiveness. In particular, Letta remains the strongest overall system under both settings, while Long-context and Embedding-RAG benefit more visibly from the stronger backbone on several query types, suggesting that generative capacity can partially compensate for weaker memory organization. By contrast, several extracted-memory methods show larger drops in the Mixed setting, which indicates that their effectiveness depends not only on memory design but also on the interaction between retrieval outputs and backbone-specific generation behavior.

\subsection{Results with Different Judge Models}
\label{sec:more-judge-models}

\begin{table*}[t]
  \centering
  \small
  \caption{Results obtained with different judge models. Each value reports the overall score (\%) under the corresponding judge model.}
  \label{tab:different-judge-models}
  \setlength{\tabcolsep}{7.5pt}
  \renewcommand{\arraystretch}{1.08}
  \resizebox{\textwidth}{!}{%
  \begin{tabular}{lcccccc}
    \toprule
    \textbf{Judge Model} & \textbf{Embedding} & \textbf{Mem0} & \textbf{A-Mem} & \textbf{Letta} & \textbf{MemRL} & \textbf{ReMem} \\
    \midrule
    claude-sonnet-4 & 42.15 & 39.20 & 44.84 & 51.21 & 46.37 & 33.93 \\
    gpt-5.1 & 40.78 & 35.03 & 44.34 & 47.69 & 42.98 & 35.07 \\
    gpt-4o & 43.30 & 38.93 & 45.12 & 54.13 & 47.20 & 35.29 \\
    gemini-2.5-pro & 46.11 & 48.32 & 49.91 & 57.75 & 50.02 & 37.79 \\
    \bottomrule
  \end{tabular}%
  }
\end{table*}

To examine whether the default judge model, Claude-sonnet-4, introduces systematic bias in evaluation, we additionally repeat all query types that require LLM-as-Judge with a different judge model. As shown in Table~\ref{tab:different-judge-models}, the overall conclusions remain largely unchanged, suggesting that the benchmark results are not sensitive to the specific judge model choice.

More specifically, the relative ordering is highly stable across judge models: Letta remains the top-performing baseline under all four judges, while ReMem stays at the bottom, and the middle-tier methods also exhibit only limited rank fluctuation. Although Gemini-2.5-Pro assigns uniformly higher absolute scores than the other judges, this shift is largely monotonic rather than selective, which suggests that changing the judge mainly affects score calibration instead of the comparative conclusions.

\subsection{Results by Chronic Disease Category}
\label{sec:more-diseases}

\begin{table*}[t]
  \centering
  \small
  \caption{Results by chronic disease category (Part I). Each value reports performance (\%) under the corresponding disease category. The best result in each column is indicated in \textbf{bold}.}
  \label{tab:results-by-disease-category-a}
  \setlength{\tabcolsep}{4.2pt}
  \renewcommand{\arraystretch}{1.08}
  \resizebox{\textwidth}{!}{%
  \begin{tabular}{lccccc}
    \toprule
    \textbf{Baselines} & \textbf{Diabetes} & \textbf{Cardiovascular} & \textbf{Chronic Respiratory} & \textbf{Neurodegenerative} & \textbf{Chronic Kidney} \\
    \midrule
    \rowcolor{blue!5}
    \multicolumn{6}{c}{\textbf{Unstructured}} \\
    BM25-RAG & 33.0 & 39.7 & 41.7 & 38.1 & 33.0 \\
    Embedding-RAG & 44.2 & 46.2 & 46.2 & 47.7 & 33.0 \\
    \midrule
    \rowcolor{blue!5}
    \multicolumn{6}{c}{\textbf{Graph-Structured}} \\
    GraphRAG & 29.9 & 46.2 & 45.4 & 47.4 & 31.4 \\
    HippoRAG-v2 & 33.5 & 33.7 & 33.7 & 37.6 & 30.5 \\
    ReMem & 31.5 & 35.2 & 29.6 & 36.0 & 29.5 \\
    \midrule
    \rowcolor{blue!5}
    \multicolumn{6}{c}{\textbf{Flat-Extracted}} \\
    Mem0 & 37.6 & 39.7 & 36.2 & 36.0 & 34.0 \\
    LightMem & 38.6 & 37.7 & 35.7 & 42.1 & 36.0 \\
    MemOS & 50.8 & 54.3 & 51.3 & 54.3 & \textbf{52.0} \\
    MemRL & 46.2 & 51.8 & 46.2 & 47.7 & 45.5 \\
    A-Mem & 46.2 & 46.2 & 43.2 & 50.8 & 43.0 \\
    \midrule
    \rowcolor{blue!5}
    \multicolumn{6}{c}{\textbf{Multi-Compartment}} \\
    Letta & \textbf{55.8} & \textbf{59.8} & \textbf{52.8} & \textbf{61.4} & 43.0 \\
    MIRIX & 38.6 & 42.7 & 38.2 & 40.1 & 33.0 \\
    \bottomrule
  \end{tabular}%
  }
\end{table*}

\begin{table*}[t]
  \centering
  \small
  \caption{Results by chronic disease category (Part II). Each value reports performance (\%) under the corresponding disease category. The best result in each column is indicated in \textbf{bold}.}
  \label{tab:results-by-disease-category-b}
  \setlength{\tabcolsep}{3.8pt}
  \renewcommand{\arraystretch}{1.08}
  \resizebox{\textwidth}{!}{%
  \begin{tabular}{lccccc}
    \toprule
    \textbf{Baselines} & \textbf{Autoimmune} & \textbf{Digestive} & \textbf{Chronic Neuropsychiatric} & \textbf{Musculoskeletal} & \textbf{Endocrine} \\
    \midrule
    \rowcolor{blue!5}
    \multicolumn{6}{c}{\textbf{Unstructured}} \\
    BM25-RAG & 36.4 & 31.7 & 33.3 & 30.2 & 42.0 \\
    Embedding-RAG & 41.4 & 41.2 & 41.9 & 40.7 & 45.0 \\
    \midrule
    \rowcolor{blue!5}
    \multicolumn{6}{c}{\textbf{Graph-Structured}} \\
    GraphRAG & 37.5 & 30.8 & 33.9 & 35.5 & 40.0 \\
    HippoRAG-v2 & 39.4 & 33.2 & 29.3 & 33.2 & 34.5 \\
    ReMem & 33.4 & 28.8 & 29.2 & 31.7 & 36.5 \\
    \midrule
    \rowcolor{blue!5}
    \multicolumn{6}{c}{\textbf{Flat-Extracted}} \\
    Mem0 & 36.8 & 38.4 & 34.4 & 37.9 & 39.5 \\
    LightMem & 37.9 & 42.2 & 33.8 & 40.7 & 43.0 \\
    MemOS & 50.0 & \textbf{51.3} & 48.5 & 53.3 & 51.0 \\
    MemRL & 52.5 & 50.3 & 46.5 & 48.2 & \textbf{56.0} \\
    A-Mem & 47.5 & 45.2 & 42.4 & 46.2 & 44.0 \\
    \midrule
    \rowcolor{blue!5}
    \multicolumn{6}{c}{\textbf{Multi-Compartment}} \\
    Letta & \textbf{55.1} & \textbf{51.3} & \textbf{55.1} & \textbf{76.5} & 55.4 \\
    MIRIX & 40.4 & 37.7 & 33.3 & 44.7 & 45.0 \\
    \bottomrule
  \end{tabular}%
  }
\end{table*}

MedMemoryBench covers 10 major chronic disease categories, and each category contains samples from two disease subtypes. To provide more medically informative analysis, we further report method performance for each chronic disease subtype. These results offer a more fine-grained view of where different methods perform well or struggle across disease settings. The detailed results are presented in Tables~\ref{tab:results-by-disease-category-a} and~\ref{tab:results-by-disease-category-b}.

The disease-wise breakdown shows that the advantage of stronger memory systems is broadly consistent rather than concentrated in a few categories. Letta achieves the best results in most disease groups, while MemOS and MemRL remain competitive in several categories such as chronic kidney, digestive, and endocrine diseases, indicating that the main findings are robust to clinical topic variation. At the same time, the gap between methods becomes especially large in categories that require integrating long-term symptoms, treatments, and temporal progression, which further supports our claim that longitudinal medical memory demands structured cross-session reasoning rather than shallow lexical matching.

\subsection{Results Across Multiple Random Seeds}
\label{sec:multiple-random-seeds}

\textbf{Experimental setup.} To assess sensitivity to randomness across repeated runs, we repeat the experiments with five random seeds: 42, 123, 456, 789, and 2024. We use GPT-5.1 as the backbone and Claude-Sonnet-4 as the judge under both the Efficient (noise-free) and Mixed (noisy) settings. Tables~\ref{tab:seed-variance-efficient} and~\ref{tab:seed-variance-mixed} report the response accuracy of each of the ten evaluated methods for every seed, together with the mean, standard deviation (Std), and coefficient of variation (CV). Accuracy is reported on a 0--1 scale, and CV is expressed as a percentage, $\mathrm{CV}=100\times\mathrm{Std}/\mathrm{Mean}$. Statistical comparisons use a significance threshold of $\alpha=0.05$, corresponding to a 95\% confidence level.

\begin{table*}[t]
  \centering
  \small
  \caption{Seed variance under the Efficient (noise-free) setting with GPT-5.1 as the backbone and Claude-Sonnet-4 as the judge. Accuracy is reported on a 0--1 scale. Mean and Std summarize five runs; CV measures relative variability. The best and second-best mean accuracies are marked in \textbf{bold} and \underline{underline}, respectively.}
  \label{tab:seed-variance-efficient}
  \setlength{\tabcolsep}{4.2pt}
  \renewcommand{\arraystretch}{1.10}
  \resizebox{\textwidth}{!}{%
  \begin{tabular}{l|ccccc!{\color{black!28}\vrule width 0.35pt}>{\columncolor{blue!7}}ccc}
    \toprule
    \textbf{Methods} & \textbf{Seed 42} & \textbf{Seed 123} & \textbf{Seed 456} & \textbf{Seed 789} & \textbf{Seed 2024} & \textbf{Mean}$\uparrow$ & \textbf{Std} & \textbf{CV (\%)} \\
    \midrule
    \rowcolor{blue!5}
    \multicolumn{9}{c}{\textbf{Graph-Structured}} \\
    GraphRAG & 0.3536 & 0.3142 & 0.3461 & 0.3155 & 0.3531 & 0.3365 & 0.0200 & 5.94 \\
    HippoRAG-v2 & 0.3539 & 0.3538 & 0.3106 & 0.3330 & 0.3333 & 0.3369 & 0.0180 & 5.34 \\
    ReMem & 0.3283 & 0.3509 & 0.3636 & 0.3285 & 0.3253 & 0.3393 & 0.0170 & 5.01 \\
    \midrule
    \rowcolor{blue!5}
    \multicolumn{9}{c}{\textbf{Flat-Extracted}} \\
    Mem0 & 0.4047 & 0.3863 & 0.3918 & 0.3701 & 0.4071 & 0.3920 & 0.0150 & 3.83 \\
    LightMem & 0.4101 & 0.3857 & 0.4056 & 0.4252 & 0.3898 & 0.4033 & 0.0160 & 3.97 \\
    MemOS & 0.4979 & 0.4748 & 0.4906 & 0.4649 & 0.4799 & \underline{0.4816} & 0.0130 & 2.70 \\
    MemRL & 0.4722 & 0.4767 & 0.4592 & 0.4395 & 0.4709 & 0.4637 & 0.0150 & 3.23 \\
    A-Mem & 0.4514 & 0.4496 & 0.4665 & 0.4471 & 0.4274 & 0.4484 & 0.0140 & 3.12 \\
    \midrule
    \rowcolor{blue!5}
    \multicolumn{9}{c}{\textbf{Multi-Compartment}} \\
    Letta & 0.5120 & 0.4967 & 0.5245 & 0.5018 & 0.5255 & \textbf{0.5121} & 0.0130 & 2.54 \\
    MIRIX & 0.3860 & 0.4208 & 0.3976 & 0.4097 & 0.3919 & 0.4012 & 0.0140 & 3.49 \\
    \bottomrule
  \end{tabular}%
  }
\end{table*}

\begin{table*}[t]
  \centering
  \small
  \caption{Seed variance under the Mixed (noisy) setting with GPT-5.1 as the backbone and Claude-Sonnet-4 as the judge. Accuracy is reported on a 0--1 scale. Mean and Std summarize five runs; CV measures relative variability. The best and second-best mean accuracies are marked in \textbf{bold} and \underline{underline}, respectively.}
  \label{tab:seed-variance-mixed}
  \setlength{\tabcolsep}{4.2pt}
  \renewcommand{\arraystretch}{1.10}
  \resizebox{\textwidth}{!}{%
  \begin{tabular}{l|ccccc!{\color{black!28}\vrule width 0.35pt}>{\columncolor{blue!7}}ccc}
    \toprule
    \textbf{Methods} & \textbf{Seed 42} & \textbf{Seed 123} & \textbf{Seed 456} & \textbf{Seed 789} & \textbf{Seed 2024} & \textbf{Mean}$\uparrow$ & \textbf{Std} & \textbf{CV (\%)} \\
    \midrule
    \rowcolor{blue!5}
    \multicolumn{9}{c}{\textbf{Graph-Structured}} \\
    GraphRAG & 0.3102 & 0.3292 & 0.3551 & 0.2816 & 0.3253 & 0.3203 & 0.0270 & 8.43 \\
    HippoRAG-v2 & 0.2857 & 0.3061 & 0.2553 & 0.3204 & 0.2805 & 0.2896 & 0.0250 & 8.63 \\
    ReMem & 0.3228 & 0.3457 & 0.3375 & 0.2898 & 0.3431 & 0.3278 & 0.0230 & 7.02 \\
    \midrule
    \rowcolor{blue!5}
    \multicolumn{9}{c}{\textbf{Flat-Extracted}} \\
    Mem0 & 0.3429 & 0.3913 & 0.3427 & 0.3538 & 0.3608 & 0.3583 & 0.0200 & 5.58 \\
    LightMem & 0.2460 & 0.2600 & 0.2975 & 0.2728 & 0.2892 & 0.2731 & 0.0210 & 7.69 \\
    MemOS & 0.4059 & 0.3997 & 0.4152 & 0.3905 & 0.4377 & \underline{0.4098} & 0.0180 & 4.39 \\
    MemRL & 0.3492 & 0.3740 & 0.3832 & 0.4023 & 0.3638 & 0.3745 & 0.0200 & 5.34 \\
    A-Mem & 0.2739 & 0.3063 & 0.2877 & 0.2721 & 0.3235 & 0.2927 & 0.0220 & 7.52 \\
    \midrule
    \rowcolor{blue!5}
    \multicolumn{9}{c}{\textbf{Multi-Compartment}} \\
    Letta & 0.4193 & 0.4131 & 0.4013 & 0.3996 & 0.4441 & \textbf{0.4155} & 0.0180 & 4.33 \\
    MIRIX & 0.2843 & 0.2804 & 0.3290 & 0.3248 & 0.3165 & 0.3070 & 0.0230 & 7.49 \\
    \bottomrule
  \end{tabular}%
  }
\end{table*}

\textbf{Statistical comparison.} We apply the Friedman test with methods as treatments and the five matched seeds as blocks. Recomputing the test from the seed-level scores of the ten methods in Tables~\ref{tab:seed-variance-efficient} and~\ref{tab:seed-variance-mixed} gives $\chi_F^2=42.47$ with 9 degrees of freedom under the Efficient setting ($p\approx2.70\times10^{-6}$), and $\chi_F^2=39.20$ with 9 degrees of freedom under the Mixed setting ($p\approx1.06\times10^{-5}$). Both asymptotic $p$-values are below 0.0001, providing evidence of overall performance differences among methods. Because the test uses only five seed blocks, its chi-squared approximation should be interpreted cautiously. Moreover, an omnibus test does not establish significance for every pair of methods or imply an identical ranking in every run.

\textbf{Cross-seed stability.} All evaluated methods have CV below 6\% in the Efficient setting and below 9\% in the Mixed setting, indicating limited relative variation across the five runs. Letta achieves the highest mean accuracy in both settings (0.5121 and 0.4155), followed by MemOS (0.4816 and 0.4098). Both also exhibit comparatively low relative variability. Nevertheless, every method has a lower mean accuracy and a higher CV in the Mixed setting than in the Efficient setting. Thus, the repeated runs support the broad performance trends while showing that noisy memory accumulation both reduces accuracy and increases sensitivity to stochastic variation. These results characterize repeatability under the evaluated configuration rather than guaranteeing invariant rankings across all seeds or experimental settings.

\subsection{Clinical Safety Evaluation}
\label{sec:clinical-safety-evaluation}

\textbf{Evaluation setup.} MedMemoryBench incorporates clinical safety evaluation through its trap-event mechanism, which tests whether agents retain and apply patient-specific constraints when answering subsequent queries. We extract 87 safety-critical queries from five patient samples (P1, P5, P10, P11, and P14) for a dedicated evaluation of Letta. These queries cover six risk categories: missed contraindications, incorrect medication recommendations, improper dosage adjustments, personalized-constraint omissions, delayed emergency responses, and unsafe over-the-counter (OTC) medication use. Table~\ref{tab:clinical-safety-letta} reports the number of correct responses out of the total queries for each category and patient. We define the safe rate as the proportion of correctly answered safety-critical queries; this benchmark-specific measure does not establish safety in clinical deployment. Two representative examples of these safety-critical scenarios are illustrated in Figure~xxx.

\begin{table*}[t]
  \centering
  \small
  \caption{Clinical safety evaluation of Letta across five patient samples. Counts denote correct responses over total safety-critical queries. Safe rate is computed from the pooled counts in each row; the overall rate is pooled across all 87 queries.}
  \label{tab:clinical-safety-letta}
  \setlength{\tabcolsep}{4.2pt}
  \renewcommand{\arraystretch}{1.10}
  \resizebox{\textwidth}{!}{%
  \begin{tabular}{l!{\color{black!22}\vrule width 0.35pt}ccccc!{\color{black!22}\vrule width 0.35pt}c>{\columncolor{blue!7}}c}
    \toprule
    \rowcolor{blue!5}
    \textbf{Risk Category} & \textbf{P1} & \textbf{P5} & \textbf{P10} & \textbf{P11} & \textbf{P14} & \textbf{Total} & \textbf{Safe Rate}$\uparrow$ \\
    \midrule
    Missed Contraindications & 2/2 & 1/2 & 2/3 & 2/3 & 1/2 & 8/12 & 66.7\% \\
    Incorrect Medication Recommendations & 4/6 & 3/5 & 2/4 & 3/4 & 3/5 & 15/24 & 62.5\% \\
    Improper Dosage Adjustments & 2/2 & 1/1 & 1/2 & 2/2 & 1/1 & 7/8 & 87.5\% \\
    Personalized-Constraint Omissions & 1/2 & 1/2 & 2/3 & 1/2 & 2/3 & 7/12 & 58.3\% \\
    Delayed Emergency Responses & 3/4 & 2/4 & 3/3 & 2/3 & 2/3 & 12/17 & 70.6\% \\
    Unsafe OTC Medication Use & 3/3 & 2/3 & 2/3 & 2/2 & 2/3 & 11/14 & 78.6\% \\
    \midrule
    \rowcolor{blue!5}
    \textbf{Overall} & \textbf{15/19} & \textbf{10/17} & \textbf{12/18} & \textbf{12/16} & \textbf{11/17} & \textbf{60/87} & \textbf{68.97\%} \\
    \bottomrule
  \end{tabular}%
  }
\end{table*}

\textbf{Results and implications.} Letta correctly answers 60 of the 87 safety-critical queries, achieving an overall safe rate of 68.97\% and leaving 27 queries incorrectly answered. Performance varies substantially across risk categories. Personalized-constraint omissions yield the lowest safe rate (58.3\%), followed by incorrect medication recommendations (62.5\%) and missed contraindications (66.7\%). These results highlight difficulties in consistently applying patient-specific information to safety-sensitive responses. Delayed emergency responses and unsafe OTC medication use attain safe rates of 70.6\% and 78.6\%, respectively, indicating that failures also occur beyond prescription-related decisions. Improper dosage adjustments achieve the highest safe rate (87.5\%), although this category contains only eight queries. Overall, this targeted evaluation reveals substantial gaps in Letta's personalized clinical safety on the evaluated samples and underscores the value of explicitly testing safety-critical memory use. Because the analysis covers only one method and five patient samples, the findings should not be interpreted as a comprehensive comparison of clinical safety across memory systems.

\subsection{Detailed Qualitative Failure Analysis}
\label{sec:qualitative-failure-analysis}

\textbf{Analysis scope.} To complement the aggregate results, we examine failure cases from A-Mem, MemOS, and Letta across the six query types in MedMemoryBench. Table~\ref{tab:failure-mode-taxonomy} summarizes the resulting taxonomy, linking each failure mode to its associated query types, observable characteristics, and the capability it tests. The reported proportions are approximate shares of the analyzed failures, not per-method error rates. Saturation degradation is treated separately as a cross-cutting phenomenon that can exacerbate other failure modes as memory accumulates. Representative failure examples are presented in Figure~xxx.

\begin{table*}[t]
  \centering
  \small
  \caption{Failure mode taxonomy from qualitative analysis of A-Mem, MemOS, and Letta. Proportions summarize the analyzed failures approximately. Saturation degradation can affect all query types and is not an additional category in the percentage breakdown.}
  \label{tab:failure-mode-taxonomy}
  \setlength{\tabcolsep}{4.2pt}
  \renewcommand{\arraystretch}{1.15}
  \begin{tabularx}{\textwidth}{@{}>{\raggedright\arraybackslash}p{2.3cm}!{\color{black!22}\vrule width 0.35pt}>{\raggedright\arraybackslash}p{2.0cm}>{\raggedright\arraybackslash}X>{\raggedright\arraybackslash}p{2.1cm}!{\color{black!22}\vrule width 0.35pt}>{\centering\arraybackslash\columncolor{blue!7}}p{1.5cm}@{}}
    \toprule
    \rowcolor{blue!5}
    \textbf{Failure Mode} & \textbf{Query Types} & \textbf{Description} & \textbf{Capability} & \textbf{Proportion} \\
    \midrule
    Retrieval Failure & EEM, TLA, MQ, IG, MCD & Key facts absent from the top-$k$ retrieved memories & Precise retrieval & $\sim$40\% \\
    Temporal Version Confusion & EEM, TLA, SUA & Historical information retrieved in place of the temporally appropriate version & Temporal indexing & $\sim$20\% \\
    Personalized Fact Omission & IG, MQ, MCD & Patient-specific trap events ignored in the response & Fact localization & $\sim$18\% \\
    Reasoning Failure & SUA, IG, MCD & Relevant evidence correctly retrieved, but subsequent reasoning diverges & Reasoning generalization & $\sim$12\% \\
    Causal Chain Breakage & IG, MCD & A required pathophysiological mechanism or causal link is missing & Multi-hop reasoning & $\sim$10\% \\
    \midrule
    \rowcolor{blue!5}
    Saturation Degradation & All & Performance degrades at later memory checkpoints & Long-horizon robustness & Cross-cutting \\
    \bottomrule
  \end{tabularx}
\end{table*}

\textbf{Retrieval, temporal alignment, and personalization.} Retrieval failure is the largest category, accounting for approximately 40\% of the analyzed failures: the evidence needed to answer a query is absent from the top-$k$ retrieved memories. Temporal version confusion accounts for approximately 20\% and reflects a different problem: a retrieved fact may be historically valid but inappropriate for the time point addressed by the query. Personalized fact omission contributes approximately 18\%, arising when responses fail to incorporate patient-specific trap events. Unlike retrieval failure, this category concerns the omission of a relevant constraint from the answer, which does not by itself establish whether the constraint was absent from the retrieved context. Together, these patterns highlight the need to retrieve not just semantically related information, but temporally appropriate and personally relevant evidence.

\textbf{Reasoning and causal-chain failures.} Reasoning failure accounts for approximately 12\% of the analyzed cases, showing that correct retrieval alone does not guarantee a correct answer. In these cases, the system accesses relevant evidence but fails to reason appropriately from it. Causal chain breakage accounts for approximately 10\% and specifically concerns missing intermediate mechanisms or links required to connect patient-specific evidence to a conclusion. Its occurrence in IG and MCD highlights the difficulty of maintaining a complete causal explanation across multiple memories, rather than merely recovering isolated facts.

\textbf{Saturation as a cross-cutting stressor.} Saturation degradation manifests as poorer performance at later evaluation checkpoints and can affect every query type. It is therefore not assigned a separate percentage alongside the five failure categories. Instead, it characterizes how failures evolve under prolonged memory accumulation, complementing the taxonomy of what goes wrong in individual responses. Taken together, the analysis motivates evaluating evidence access, temporal alignment, personalized constraint use, and downstream reasoning jointly, while tracking their robustness over time. These qualitative findings characterize the inspected cases and do not establish the relative prevalence of each failure mode for every method or for the benchmark as a whole.

\section{Prompt Template}
\label{sec:prompt-template}

\subsection{Evaluation Prompt Used During Benchmarking}
\label{sec:evaluation-prompt}

MedMemoryBench uses a two-stage prompting pipeline during benchmarking: \textbf{Memory Construction} and \textbf{Query Answering}. In the prompt templates below, placeholders are written as \texttt{<field\_name>} and are filled dynamically at runtime. For example, \texttt{<context>} refers to the current medical dialogue record, \texttt{<memory\_source>} denotes the retrieved or accumulated memory used for answering, and \texttt{<question>} refers to the benchmark query.

{%
\tcbset{
  medprompt/.style={
    enhanced,
    breakable,
    rounded corners,
    arc=2.4mm,
    boxrule=0.8pt,
    colframe=blue!55!violet,
    colback=black!0.8,
    coltitle=violet!65!black,
    fonttitle=\bfseries,
    left=2.0mm,
    right=2.0mm,
    top=1.8mm,
    bottom=1.6mm,
    before skip=0.7em,
    after skip=0.9em,
    valign=top,
    attach boxed title to top left={xshift=2.2mm,yshift*=-2.0mm},
    boxed title style={
      rounded corners,
      arc=1.8mm,
      boxrule=0.75pt,
      colframe=blue!45!violet,
      colback=black!2,
      left=1.4mm,
      right=1.4mm,
      top=0.8mm,
      bottom=0.8mm
    }
  },
  judgeprompt/.style={
    enhanced,
    breakable,
    rounded corners,
    arc=1.2mm,
    boxrule=0.95pt,
    colframe=violet!55!blue!55,
    colback=blue!2!violet!1,
    coltitle=violet!45!black,
    fonttitle=\bfseries,
    left=2.0mm,
    right=2.0mm,
    top=1.6mm,
    bottom=1.4mm,
    before skip=0.7em,
    after skip=0.9em,
    valign=top,
    attach boxed title to top left={xshift=2.0mm,yshift*=-2.0mm},
    boxed title style={
      rounded corners,
      arc=1.0mm,
      boxrule=0.8pt,
      colframe=violet!48!blue!50,
      colback=blue!8!violet!6,
      left=1.3mm,
      right=1.3mm,
      top=0.7mm,
      bottom=0.7mm
    }
  }
}
\newcommand{\promptph}[1]{\textcolor{violet!75!blue}{\texttt{<#1>}}}
\newcommand{\promptctx}[1]{\textcolor{blue!65!black}{\texttt{<#1>}}}
\newcommand{\promptans}[1]{\textcolor{teal!55!black}{\texttt{<#1>}}}
\newcommand{\promptmeta}[1]{\textcolor{orange!80!black}{\texttt{<#1>}}}
\newcommand{\promptjudge}[1]{\textcolor{red!65!black}{\texttt{<#1>}}}
\newcommand{\promptlabel}[1]{\textcolor{black!82}{\textbf{#1}}}
\newcommand{\promptreq}[1]{\textcolor{violet!70!black}{\textbf{#1}}}
\newcommand{\promptcrit}[1]{\textcolor{blue!55!black}{\textbf{#1}}}
\newcommand{\promptwarn}[1]{\textcolor{red!65!black}{\textbf{#1}}}
\newcommand{\promptgood}[1]{\textcolor{teal!60!black}{\textbf{#1}}}
\newcommand{\promptjson}[1]{\textcolor{violet!60!black}{\texttt{#1}}}

\paragraph{Memory Construction.} In the memory-construction stage, the model receives the current dialogue context and is asked to read and memorize the medically salient information. We use one shared template for this stage.

\begin{tcolorbox}[medprompt,title=Memorize Prompt]
\ttfamily\small
\textbf{Instruction:} The following is a \textbf{medical dialogue record}. Please read it carefully and \textbf{memorize the key information}.\par
\medskip
\promptctx{context}
\end{tcolorbox}

This prompt is intentionally lightweight. Its purpose is to encourage faithful retention of clinically important facts from the dialogue record without adding unnecessary reasoning constraints during memory writing.

\paragraph{Query Answering.} In the query-answering stage, we first provide a shared system instruction to define the assistant role, and then pair it with a query-specific user prompt template. Because MedMemoryBench contains heterogeneous question types with different answer targets and reasoning demands, we use a dedicated template for each query category.

\begin{tcolorbox}[medprompt,title=Shared System Prompt]
\ttfamily\small
You are the patient's \textbf{personalized medical assistant}, capable of accurately memorizing the patient's complete medical history. Please \textbf{reason and respond based on patient information in memory}, maintain a warm yet professional tone, answer directly, and avoid unnecessarily long explanations.
\end{tcolorbox}

\begin{tcolorbox}[medprompt,title=EEM Answer Prompt]
\ttfamily\small
\promptlabel{Context:} Based on \promptctx{memory\_source}, accurately answer the following question.\par
\medskip
\promptlabel{Question:} \promptctx{question}\par
\medskip
\promptreq{Answer Requirements:}\par
\textbf{1.} Provide the target entity name directly.\par
\textbf{2.} Keep the answer \textbf{brief and precise}.\par
\textbf{3.} Do not include lengthy explanations.\par
\medskip
\promptlabel{Answer:}
\end{tcolorbox}

The \textbf{EEM} template is designed for precise slot-level retrieval. It therefore emphasizes direct extraction of the target entity rather than extended explanation.

\begin{tcolorbox}[medprompt,title=TLA Answer Prompt]
\ttfamily\small
\promptlabel{Context:} Based on \promptctx{memory\_source}, accurately answer the following question.\par
\medskip
\promptlabel{Question:} \promptctx{question}\par
\medskip
\promptreq{Answer Requirements:}\par
\textbf{1.} If the question asks about a time, answer in \textbf{YYYY-MM-DD} format (e.g., \texttt{2024-01-15}).\par
\textbf{2.} If the question asks about an event at a specific time, clearly describe the \textbf{event content and key details}.\par
\textbf{3.} Keep the answer concise and directly grounded in memory.\par
\medskip
\promptlabel{Answer:}
\end{tcolorbox}

The \textbf{TLA} template explicitly constrains temporal questions to normalized date outputs whenever possible, while still allowing concise event descriptions when the query asks what happened at a particular time point.

\begin{tcolorbox}[medprompt,title=SUA Answer Prompt]
\ttfamily\small
\promptlabel{Context:} Based on \promptctx{memory\_source}, accurately answer the following question.\par
\medskip
\promptlabel{Question:} \promptctx{question}\par
\medskip
\promptreq{Answer Requirements:}\par
\textbf{1.} Describe the patient's \textbf{most recent status}.\par
\textbf{2.} Reflect important \textbf{changes over time} when necessary.\par
\textbf{3.} Maintain a \textbf{warm yet professional tone}.\par
\textbf{4.} Be concise and direct.\par
\medskip
\promptlabel{Answer:}
\end{tcolorbox}

The \textbf{SUA} prompt emphasizes up-to-date patient status and trajectory-aware summarization, which is important for questions that ask for the latest condition rather than isolated historical facts.

\begin{tcolorbox}[medprompt,title=MQ Answer Prompt]
\ttfamily\small
\promptlabel{Context:} Based on \promptctx{memory\_source}, and considering the patient's allergy history, medical history, medications, and personal preferences, answer the following question.\par
\medskip
\promptlabel{Question:} \promptctx{question}\par
\medskip
\promptreq{Answer Requirements:}\par
\textbf{1.} Select \textbf{all} correct options.\par
\textbf{2.} Output \textbf{only} the option letter(s), such as \texttt{B} or \texttt{B,D}.\par
\textbf{3.} Do \textbf{not} provide any explanation.\par
\medskip
\promptlabel{Answer:}
\end{tcolorbox}

The \textbf{MQ} template enforces a strict multiple-choice output format, which simplifies automatic evaluation and avoids verbose justifications that are irrelevant to the benchmark target.

\begin{tcolorbox}[medprompt,title=IG Answer Prompt]
\ttfamily\small
\promptlabel{Context:} Based on \promptctx{memory\_source}, and considering the patient's allergy history, medical history, medications, and personal preferences, answer the following question.\par
\medskip
\promptlabel{Question:} \promptctx{question}\par
\medskip
\promptreq{Answer Requirements:}\par
\textbf{1.} Reason from this patient's \textbf{specific remembered information}; do not give generic medical advice.\par
\textbf{2.} Maintain a \textbf{warm yet professional tone}.\par
\textbf{3.} Be \textbf{concise, direct}, and avoid boilerplate.\par
\textbf{4.} If recommending or advising against something, briefly explain the reason based on the patient's specific situation.\par
\medskip
\promptlabel{Answer:}
\end{tcolorbox}

The \textbf{IG} template is designed for personalized medical inference. It explicitly discourages generic recommendations and instead requires patient-grounded reasoning based on remembered allergy history, medication use, prior diagnoses, and personal preferences.

\begin{tcolorbox}[medprompt,title=MCD Answer Prompt]
\ttfamily\small
\promptlabel{Context:} Based on \promptctx{memory\_source}, carefully review the patient's complete medical history and conduct a comprehensive analysis by combining information from multiple visits.\par
\medskip
\promptlabel{Question:} \promptctx{question}\par
\medskip
\promptreq{Answer Requirements:}\par
\textbf{1.} Clearly list the \textbf{memory content} you draw upon.\par
\textbf{2.} Present a \textbf{clear reasoning path} from evidence to conclusions.\par
\textbf{3.} Provide a \textbf{final comprehensive judgment}.\par
\medskip
\promptlabel{Answer:}
\end{tcolorbox}

Finally, the \textbf{MCD} template targets queries that require multi-visit synthesis and explicit reasoning over dispersed historical evidence. Compared with the other templates, it places the strongest emphasis on transparent reasoning paths and comprehensive judgment grounded in multiple memory items.

\subsection{LLM-as-Judge Prompt Template}
\label{sec:judge-prompt}

For MedMemoryBench, we use LLM-as-Judge for four query types whose outputs cannot be reliably evaluated by exact string matching alone: \textbf{TLA}, \textbf{SUA}, \textbf{IG}, and \textbf{MCD}. These categories require judgment over temporal equivalence, state consistency, patient-grounded inference, or multi-hop reasoning quality. Accordingly, we design a dedicated judge prompt for each type so that the evaluator can apply query-specific criteria rather than a uniform matching rule.

More specifically, \textbf{TLA} requires verifying whether the model correctly identifies a time point or correctly describes the event associated with a queried time; \textbf{SUA} requires checking whether the answer reflects the patient's \emph{latest} status and whether it is grounded in remembered patient history rather than generic knowledge; \textbf{IG} requires evaluating whether the model reaches the right personalized recommendation or conclusion by explicitly using patient-specific evidence; and \textbf{MCD} requires a substantially stricter assessment of node coverage, causal-chain correctness, and reasoning completeness across multiple visits.

Below we present the prompt templates used for LLM-based judgment.

\begin{tcolorbox}[judgeprompt,title=TLA Judge Prompt]
\ttfamily\small
You are a strict medical dialogue evaluation judge. Determine whether the model's answer correctly addresses the time-related question.\par
\medskip
\promptlabel{Question:}\par
\promptctx{question}\par
\medskip
\promptgood{Reference Answer:}\par
\promptans{expected\_answer}\par
\medskip
\promptreq{Answer Explanation:}\par
\promptmeta{explanation}\par
\medskip
\promptwarn{Model's Answer:}\par
\promptjudge{model\_output}\par
\medskip
\textbf{Evaluation Criteria:}\par
This is a temporal localization question, which may take one of the following two forms:\par
1. Asking when a certain event occurred. The model must correctly provide the time point\par
2. Asking what happened at a certain time. The model must correctly describe the event content\par
\medskip
Judge strictly:\par
- If the model's answer contains the correct time point or the correct event content, judge as [CORRECT]\par
- If the model's answer about the time/event does not match the reference answer or fails to answer, judge as [INCORRECT]\par
- Date formats do not need to be identical, but must refer to the same time point (e.g., ``January 1, 2024'' and ``2024-01-01'' are considered equivalent)\par
\medskip
Output in the following JSON format:\par
\promptjson{{"is\_correct": true/false, "reason": "brief justification"}}\par
\medskip
Output JSON only, no other content.
\end{tcolorbox}

The \textbf{TLA} judge prompt focuses on temporal equivalence. It allows surface-form variation in date expressions, but requires exact agreement at the level of the referenced time point or event content.

\begin{tcolorbox}[judgeprompt,title=SUA Judge Prompt]
\ttfamily\small
You are a very strict medical dialogue evaluation judge. Determine whether the model's answer correctly reflects the patient's most recent status.\par
\medskip
\promptlabel{Question:}\par
\promptctx{question}\par
\medskip
\promptgood{Reference Answer:}\par
\promptans{expected\_answer}\par
\medskip
\promptreq{Answer Explanation:}\par
\promptmeta{explanation}\par
\medskip
\promptwarn{Model's Answer:}\par
\promptjudge{model\_output}\par
\medskip
\textbf{Evaluation Criteria:}\par
This is a state update question, testing whether the model correctly answers the latest status based on the patient's historical information in memory.\par
\medskip
Core Evaluation Principles (critically important):\par
1. \textbf{Must be based on memory}: The model's answer must demonstrate the use of the patient's past memory information, not guessing or generic medical knowledge.\par
2. \textbf{No guessing allowed}: If the model has not retrieved relevant memory information but gives a ``coincidentally correct'' answer, it should be judged as [INCORRECT].\par
3. \textbf{Information source requirement}: A correct answer should convey that the model ``remembers'' this patient's specific situation, rather than guessing.\par
\medskip
Judge strictly:\par
- If the model's answer demonstrates the use of patient historical memory and the core content is consistent with the reference answer, judge as [CORRECT]\par
- If the model's answer contains key information points from the reference answer, and these clearly originate from patient memory retrieval, judge as [CORRECT]\par
- If the model's answer clearly contradicts the reference answer, omits key information, or provides outdated status, judge as [INCORRECT]\par
- If the model states it does not know or cannot answer, judge as [INCORRECT]\par
- If the model's answer appears too generic, lacks specific patient information support, or seems like a guess, even if the content happens to be close to the reference answer, judge as [INCORRECT]\par
\medskip
Output in the following JSON format:\par
\promptjson{{"is\_correct": true/false, "reason": "brief justification, must indicate whether the model demonstrated use of patient memory"}}\par
\medskip
Output JSON only, no other content.
\end{tcolorbox}

The \textbf{SUA} judge prompt is designed to distinguish genuine memory-grounded answers from plausible but unsupported status summaries. In particular, it explicitly treats unsupported guessing as incorrect even when the final answer is superficially close to the reference.

\begin{tcolorbox}[judgeprompt,title=IG Judge Prompt]
\ttfamily\small
You are a very strict medical dialogue evaluation judge. Determine whether the model's reasoning answer is correct.\par
\medskip
\promptlabel{Question:}\par
\promptctx{question}\par
\medskip
\promptgood{Reference Answer:}\par
\promptans{expected\_answer}\par
\medskip
\promptreq{Answer Explanation:}\par
\promptmeta{explanation}\par
\promptmeta{metadata\_info}\par
\medskip
\promptwarn{Model's Answer:}\par
\promptjudge{model\_output}\par
\medskip
\textbf{Evaluation Criteria:}\par
This is an inference generation question, testing whether the model can perform correct medical reasoning based on patient-specific information.\par
\medskip
Core evaluation points:\par
\medskip
1. \promptreq{Patient Information Utilization (Key)}\par
- The model must demonstrate the use of patient-specific information from memory\par
- If \texttt{required\_patient\_info} is provided in metadata, the model's answer must reflect understanding of these key pieces of information (important)\par
- If the patient's specific circumstances and past memories are ignored or missing, judge as [INCORRECT]\par
\medskip
2. \promptreq{Reasoning Quality}\par
- The model must reason based on retrieved patient historical information, not purely from its own medical common sense\par
- If only a conclusion is given without sufficient reference to patient information and memory, judge as [INCORRECT]\par
- If the model gives a ``common wrong answer'' type of response (generic advice), judge as [INCORRECT]\par
\medskip
3. \promptreq{Conclusion Correctness}\par
- The final recommendation/conclusion should be fully consistent with the reference answer in direction\par
- Even if the conclusion is correct, if it lacks reasoning based on patient information, still judge as [INCORRECT]\par
\medskip
Judgment rules:\par
- [CORRECT]: Answer uses patient-specific information, contains required patient information points, and reaches an accurate conclusion\par
- [INCORRECT]: Answer does not adequately consider the patient's specific circumstances\par
- [INCORRECT]: Answer ignores certain key information in \texttt{required\_patient\_info}\par
- [INCORRECT]: Answer matches the \texttt{common\_wrong\_answer} pattern\par
- [INCORRECT]: Model refuses to answer or claims no information\par
\medskip
Output in the following JSON format:\par
\promptjson{{"is\_correct": true/false, "reason": "brief justification"}}\par
\medskip
Output JSON only, no other content.
\end{tcolorbox}

The \textbf{IG} judge prompt emphasizes patient-grounded reasoning rather than generic recommendation quality. This design is especially important because clinically plausible but non-personalized advice should not receive credit in MedMemoryBench.

\begin{tcolorbox}[judgeprompt,title=MCD Judge Prompt]
\ttfamily\small
You are an \textbf{extremely strict} medical multi-hop reasoning evaluation judge. Your task is to rigorously verify whether the model truly retrieved and used specific information from the patient's historical memory to perform multi-hop clinical reasoning.\par
\medskip
\promptlabel{Question:}\par
\promptctx{question}\par
\medskip
\promptgood{Reference Answer:}\par
\promptans{expected\_answer}\par
\medskip
\promptreq{Answer Explanation:}\par
\promptmeta{explanation}\par
\promptmeta{nodes\_for\_validation}\par
\promptmeta{required\_nodes\_str}\par
\promptreq{Reasoning Hops:} \promptmeta{hop\_count}\par
\promptreq{Reasoning Pattern:} \promptmeta{reasoning\_pattern}\par
\medskip
\promptwarn{Model's Answer:}\par
\promptjudge{model\_output}\par
\medskip
---\par
\medskip
\promptcrit{Evaluation Task: Strict Node-by-Node Reasoning Chain Verification}\par
\medskip
This is a multi-hop clinical reasoning question. \textbf{The core assessment is whether the model can accurately retrieve and use specific personalized medical information from the patient's historical memory}.\par
\medskip
\promptcrit{Key Evaluation Principles (must be strictly followed)}\par
\medskip
1. \promptreq{Patient-Specific Information Principle:} The model must explicitly reference the patient's \textbf{specific data} (such as specific test values, medication dosages, specific timing of symptom onset, particular diagnostic results), rather than giving generic medical common sense.\par
- ``Poor blood sugar control may lead to...''. This is generic medical knowledge, not patient-specific information\par
- ``Your fasting blood glucose rose from 6.8 to 8.2...''. This is patient-specific information\par
\medskip
2. \promptreq{Memory Retrieval Evidence Principle:} If the model fails to demonstrate specific references to the patient's historical records, even if the reasoning direction is correct, it should be judged as \textbf{inadequate}. The model must show it ``remembers'' the patient's specific situation.\par
\medskip
3. \promptreq{Strict Causal Chain Correspondence Principle:} The causal relationships established by the model must \textbf{precisely correspond} to the causal mechanisms described in the reasoning chain nodes. Similar but different mechanisms cannot substitute.\par
\medskip
4. \promptreq{Node Content Precise Matching Principle:} During node verification, it is not sufficient to judge as ``covered'' merely because the model mentioned a related concept. You must verify whether the model referenced the \textbf{core specific content} within the node.\par
\medskip
\promptcrit{Evaluation Steps}\par
\medskip
\promptcrit{Step 1: Strict Node-by-Node Check}\par
For each node in the reasoning chain, all of the following conditions must be verified:\par
\medskip
\promptreq{Condition A - Specific Information Match:}\par
- Did the model mention the \textbf{specific data/time/event} in this node?\par
- If the node contains specific values (e.g., ``TSH 0.02''), the model must mention the same or equivalent value\par
- If the node contains a specific time (e.g., ``October 2024''), the model must demonstrate awareness of that time point\par
- Merely mentioning the related concept (e.g., ``thyroid function'') without specific data \textbf{does not count as coverage}\par
\medskip
\promptreq{Condition B - Correct Causal Mechanism:}\par
- Does the causal mechanism described by the model \textbf{exactly match} the reference reasoning chain?\par
- Using a different pathophysiological explanation (even if it sounds reasonable) \textbf{does not count as correct}\par
- Skipping intermediate steps to reach a conclusion directly \textbf{does not count as correct}\par
\medskip
\promptreq{Condition C - Clear Information Source:}\par
- Does the model's answer clearly demonstrate that this information comes from the patient's historical memory?\par
- Inferences based purely on medical common sense \textbf{cannot receive credit}\par
\medskip
\promptcrit{Step 2: Calculate Three Scoring Dimensions (Strict Standards)}\par
\medskip
\promptgood{NCR (Node Coverage Rate)} = Number of nodes fully satisfying Condition A / Total number of nodes\par
- Mentioning concept only without specific data. Node not counted as covered\par
- Incorrect data or mismatched timeline. Node not counted as covered\par
\medskip
\promptgood{CRC (Causal Relation Correctness)} = Number of correctly established causal links / Number of expected causal links\par
- Must be the causal mechanism described in the reference answer; ``equivalent substitutions'' not accepted\par
- Causal links skipping intermediate nodes. No credit\par
\medskip
\promptgood{CC (Chain Completeness)}\par
- 1.0 = Complete coverage of all nodes with correct causal relations\par
- 0.7 = Coverage of 80\%+ nodes, core causal relations correct\par
- 0.5 = Coverage of 60\%+ nodes, main causal relations correct\par
- 0.3 = Partial node coverage, causal relations have gaps\par
- 0.0 = No valid reasoning chain or completely wrong direction\par
\medskip
\promptcrit{Step 3: Comprehensive Judgment (High Standards)}\par
\medskip
\promptgood{[CORRECT] Conditions (all must be satisfied simultaneously):}\par
- NCR $\geq$ 0.75 (at least three-quarters of nodes' specific content covered)\par
- CRC $\geq$ 0.75 (causal relations basically complete and correct)\par
- CC $\geq$ 0.7 (reasoning chain basically complete)\par
- Final conclusion consistent with reference answer\par
\medskip
\promptreq{[PARTIALLY CORRECT] Conditions:}\par
- NCR $\geq$ 0.5 and CRC $\geq$ 0.5 and CC $\geq$ 0.5\par
- Main reasoning direction correct, but with notable gaps\par
\medskip
\promptwarn{[INCORRECT] Conditions (any one triggers incorrect judgment):}\par
- Model failed to retrieve patient-specific information from memory\par
- Reasoning based on generic medical knowledge rather than patient-specific situation\par
- Causal mechanism inconsistent with reference answer\par
- Conclusion direction incorrect\par
- NCR $<$ 0.5 or CRC $<$ 0.5 or CC $<$ 0.5\par
\medskip
\promptcrit{Output Format}\par
\medskip
Output your strict evaluation result in the following JSON format:\par
\{\par
\quad ``node\_validations'': [\par
\quad\quad \{\par
\quad\quad\quad ``node\_id'': 1,\par
\quad\quad\quad ``mentioned'': true/false,\par
\quad\quad\quad ``specific\_data\_matched'': true/false,\par
\quad\quad\quad ``causal\_link\_correct'': true/false,\par
\quad\quad\quad ``note'': ``Must state: 1) What specific data the model mentioned 2) Whether it precisely matches node content 3) Whether the causal relation is correct''\par
\quad\quad \}\par
\quad ],\par
\quad ``ncr\_score'': 0.0-1.0,\par
\quad ``crc\_score'': 0.0-1.0,\par
\quad ``cc\_score'': 0.0-1.0,\par
\quad ``memory\_retrieval\_quality'': ``excellent/good/partial/poor/none'',\par
\quad ``uses\_patient\_specific\_info'': true/false,\par
\quad ``is\_correct'': true/false,\par
\quad ``reason'': ``Comprehensive justification, must state: 1) Whether the model used patient-specific information 2) Which nodes were not covered 3) Whether the causal chain is complete''\par
\}\par
\medskip
Output JSON only, no other content.
\end{tcolorbox}

Among the four judge prompts, the \textbf{MCD} prompt is the most stringent. Instead of evaluating only the final answer, it requires explicit verification of patient-specific evidence usage, intermediate node coverage, and causal-chain faithfulness. This design makes the judge more sensitive to whether a model truly performs multi-hop reasoning over longitudinal medical memory rather than producing a plausible high-level summary.

}

\section{Special Implementation Details of the Baselines}
\label{sec:special-implementation}

For baseline implementation within the MedMemoryBench evaluation framework, we re-engineered most methods' HTTP clients, timeout handling, and TCP keepalive configurations to support more robust large-scale evaluation and accurate accounting of API token consumption. For several baselines, additional method-specific adaptations were necessary due to constraints in their original implementations.

\subsection{MIRIX}
\label{sec:implementation-mirix}

In the original MIRIX framework, the interactive dialogue mode relies on an internal MetaAgent to decide the retrieval strategy automatically. To ensure fair evaluation and precise token accounting, we instead use a Manual mode: we explicitly call \texttt{retrieve\_memory(memory\_type="all")} to perform fused retrieval over all five memory types, namely Episodic, Semantic, Procedural, Resource, and Knowledge Vault, and then use an external LLM to generate the final answer. Whereas the original MIRIX retrieval is designed for a single memory type at a time, this all-type fused retrieval is a task-specific adaptation introduced for our evaluation framework.

\subsection{HippoRAG-v2}
\label{sec:implementation-hipporag}

HippoRAG-v2 is originally designed for one-shot indexing of a document collection followed by querying. Applying this design directly in our setting would incur substantial time overhead. We therefore implement a cumulative-trigger scheme: for non-final sessions, raw text is only buffered into \texttt{\_pending\_sessions}; when \texttt{is\_last\_session=True}, we invoke \texttt{hipporag.index(docs)} once to build the knowledge graph in batch.

\subsection{ReMem}
\label{sec:implementation-remem}

Because each session in MedMemoryBench is relatively long, ReMem's internal chunking pipeline would introduce excessive API calls, leading to substantial time and token costs. We therefore set \texttt{preprocess\_chunk\_func="none"} to disable ReMem's internal text chunking. Each session is passed as a single document, and most sessions are thus processed without further splitting, substantially reducing the number of LLM calls.

\section{Ethical, Privacy, and Safety Considerations}
\label{sec:ethics-privacy-safety}

\textbf{Ethical review scope.} MedMemoryBench is constructed for research on long-term medical memory rather than for clinical deployment. The benchmark data are generated and curated within a controlled pipeline and do not contain directly identifiable personal information. Even so, we treated the benchmark as a medical-domain resource with elevated risk and conducted internal review of data generation, annotation, storage, and release procedures before using it in experiments.

\textbf{Privacy protection.} Our construction pipeline is designed to avoid exposing personal identifiers at the source. During annotation and quality control, we restrict access to authorized staff, avoid recording unnecessary personal attributes, and focus review on medical consistency, answer correctness, and expression standardization. We do not release annotator identities, internal annotation records, or other operational metadata that could create secondary privacy risks.

\textbf{Risk mitigation and intended use.} We position MedMemoryBench as a research-only benchmark for evaluating memory retrieval and reasoning under longitudinal medical dialogue. It should not be used to support diagnosis, treatment recommendation, triage, or other real clinical decisions. We therefore recommend that future users keep human oversight in the loop, avoid deploying benchmark prompts or outputs directly in patient-facing systems, and perform separate ethics, privacy, and safety review if they extend the benchmark with real clinical data. Any future expansion toward real-world or multimodal medical records should additionally undergo institution-specific ethical review, stricter de-identification, and access control before public release.

\phantomsection
\hypertarget{app:limitation-future-work}{}
\section{Limitation and Future Work}
\label{sec:limitation-future-work}

\textbf{Synthetic-data bias remains unavoidable.} Although we introduced human annotation for verification and correction, the raw dialogues and much of the underlying content are still generated by LLMs. As a result, the benchmark may retain model-specific priors and may not fully match the diversity of real patient--doctor interactions.

\textbf{Scale and coverage are still limited.} Due to annotation cost and budget constraints, the disease spectrum is not exhaustive and may be biased toward conditions that are easier to synthesize and standardize. In addition, despite its relatively large size, the dataset still contains limited numbers of sessions and patient trajectories.

\textbf{The evaluation setting has limited ecological validity.} Our benchmark mainly evaluates memory retrieval and reasoning in a controlled offline setting. It does not explicitly assess memory interpretability or editability, nor does it provide a deep evaluation of privacy and safety, which are important in medical applications.

\textbf{The benchmark is text-only.} MedMemoryBench does not yet cover multimodal medical memory, such as laboratory reports, medical images, prescriptions, or wearable signals, and therefore cannot fully reflect real-world clinical memory scenarios.

For future work, we plan to extend the benchmark toward multimodal medical data and to incorporate real user interaction data, so that the benchmark can better capture practical healthcare workflows and more realistic memory evolution patterns.


\end{document}